\documentclass{article}

\usepackage{graphicx} 
\usepackage{float}
\usepackage{authblk}

\usepackage{algorithm}
\usepackage{algorithmic}
\usepackage{multirow}
\usepackage{booktabs}
\usepackage{float}
\usepackage{makecell}
\usepackage{makecell}
\usepackage{CJKutf8}
\usepackage[numbers,sort&compress]{natbib}
\usepackage{enumitem}

\usepackage[colorlinks=true, urlcolor=blue, linkcolor=red]{hyperref}

\title{Revolutionizing Finance with LLMs: An Overview of Applications and Insights}

\author{Huaqin Zhao, Zhengliang Liu, Zihao Wu, Yiwei Li, Tianze Yang, Peng Shu, Shaochen Xu, Haixing Dai, Lin Zhao, Hanqi Jiang, Yi Pan, Junhao Chen, Yifan Zhou, Wei Ruan, Zheyuan Zhang, Zeyu Zhang, Ruitong Sun, Gengchen Mai, Ninghao Liu, Tianming Liu$\dagger$\thanks{$\dagger$Corresponding authors: Tianming Liu}\thanks{Huanqin Zhao, Zhengliang Liu, Zihao Wu, Yiwei Li, Tianze Yang, Peng Shu, Shaochen Xu, Haixing Dai, Lin Zhao, Hanqi Jiang, Yi Pan, Junhao Chen, Yifan Zhou, Wei Ruan, Ruitong Sun, Ninghao Liu, Tianming are with the School of Computing, The University of Georgia, Athens 30602, USA. Zheyuan Zhang is with the Northwestern University. Zeyu Zhang is with the Computer Science and Engineering, The University of Texas at Arlington, Arlington, TX, USA. Gengchen Mai is with the Department of Geography, University of Georgia, Athens 30602, USA.}}

\begin{document}

\maketitle

\begin{abstract}
In recent years, Large Language Models (LLMs) like ChatGPT have seen considerable advancements and have been applied in diverse fields. Built on the Transformer architecture, these models are trained on extensive datasets, enabling them to understand and generate human language effectively. In the financial domain, the deployment of LLMs is gaining momentum. These models are being utilized for automating financial report generation, forecasting market trends, analyzing investor sentiment, and offering personalized financial advice. Leveraging their natural language processing capabilities, LLMs can distill key insights from vast financial data, aiding institutions in making informed investment choices and enhancing both operational efficiency and customer satisfaction. 
In this study, we provide a comprehensive overview of the emerging integration of LLMs into various financial tasks. Additionally, we conducted holistic tests on multiple financial tasks through the combination of natural language instructions. Our findings show that GPT-4 effectively follow prompt instructions across various financial tasks.
This survey and evaluation of LLMs in the financial domain aim to deepen the understanding of LLMs' current role in finance for both financial practitioners and LLM researchers, identify new research and application prospects, and highlight how these technologies can be leveraged to solve practical challenges in the finance industry.
\end{abstract}

\begin{figure*}[t]
\centering
\includegraphics[width=1\textwidth]{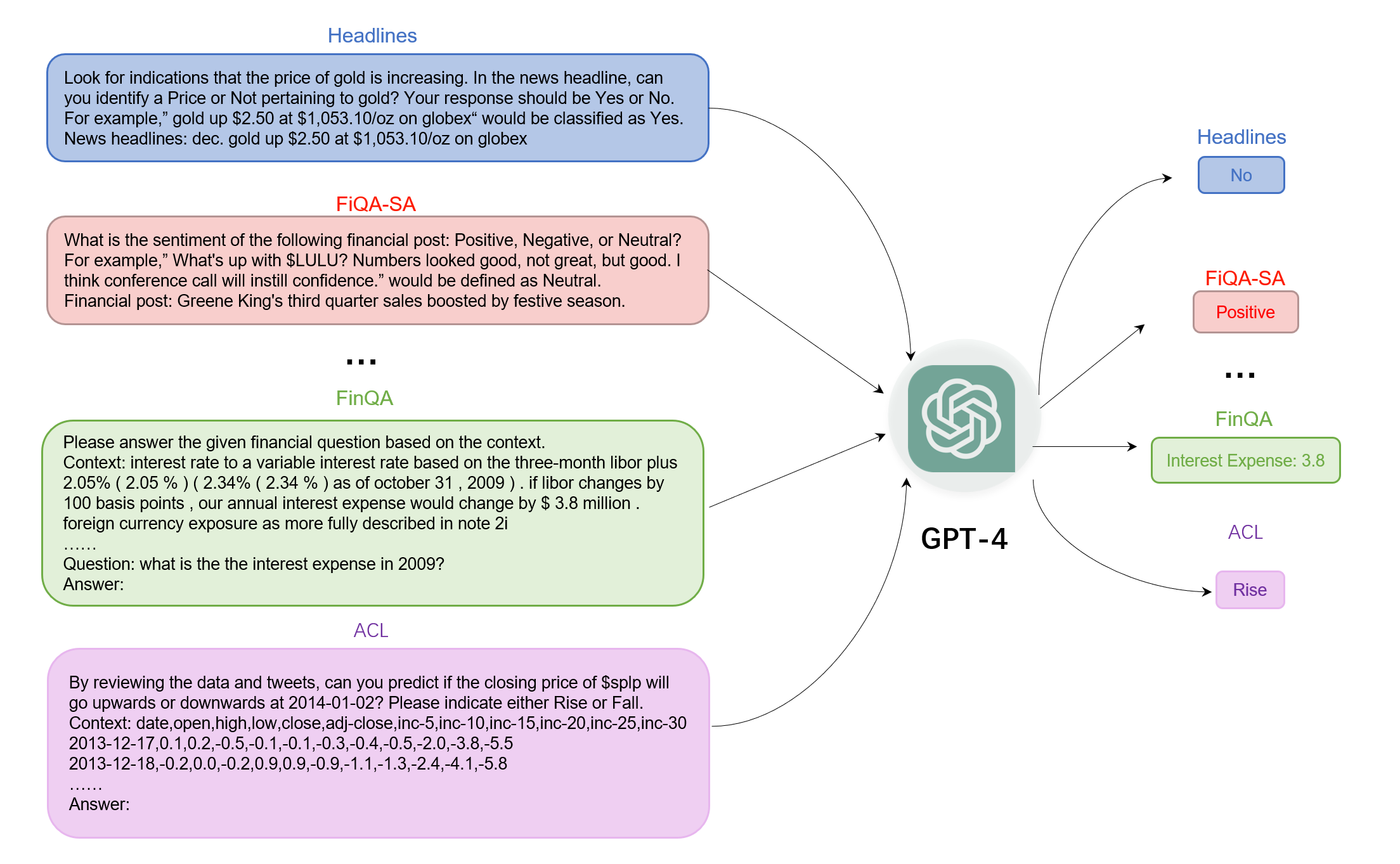}
\caption{An overview of the LLMs' capacities in financial tasks.} 
\end{figure*}

\section{Introduction}
Over the past few years, LLMs such as OpenAI's GPT family have made significant advances in the field of natural language processing (NLP). The development of these models marks an important milestone in AI technology for understanding and generating natural language. With increased computational power and improved algorithms, LLMs has demonstrated amazing capabilities in understanding complex contexts, answering questions, and writing content. Especially in the finance domain, these capabilities of LLMs are gradually showing their great potential \cite{chang2023survey,wu2023brief,wu2023bloomberggpt}.

Finance is a highly specialized and complex field that involves a great deal of data analysis, prediction, and decision making. LLM's ability to process large-scale text data makes it a promising application in the financial field. For example, by analyzing financial reports, market news, investor communications, etc., LLMs can provide insights into market trends, perform risk assessments, and even assist in investment decisions. In addition, LLMs can process natural language queries and provide instant financial advice and support, which is a big step forward for the financial services industry \cite{wu2023bloomberggpt,liu2023fingpt,deng2023llms}.

However, applying LLMs to the financial sector also faces several challenges. First, data in the financial domain is highly specialized and complex. Financial terminology, regulations, and market dynamics require a high level of model comprehension. In addition, financial decision-making usually involves high risk, which requires a high degree of accuracy and reliability in prediction. Therefore, it is a major challenge to ensure that the output of LLMs is both accurate and reliable \cite{gupta2023gpt}.

To address these issues, researchers and developers are continuously refining the algorithms of LLMs to improve its understanding and processing of specialized domain knowledge. With a large amount of specialized training data, the model can better grasp specific knowledge in the financial domain. At the same time, the combination of expert systems and manual review mechanisms can further improve the accuracy and reliability of the model's application in the financial domain \cite{yang2020finbert}.

Overall, large-scale language models are gradually becoming a powerful tool for dealing with financial problems. They are not only able to process and analyze large amounts of data, but also provide in-depth insights and recommendations \cite{wu2023bloomberggpt}. Although there are some challenges in the application process, they are gradually being overcome as technology continues to advance. Looking ahead, the application of LLMs in the financial sector will undoubtedly open up more innovations and opportunities.

In this review, we tackle the key question of how to address the difficulties inherent in the financial sector while utilizing the successes of LLMs from various fields to enhance the finance industry. The significant contributions of this article are distilled into four primary points, each focusing on the synergy between LLMs and financial applications.
\begin{itemize}
\item We meticulously survey and synthesize existing LLMs for finance literature, exploring the latest advancements in four independent task categories: financial engineering, financial forecasting, financial risk management, and financial real-time question answering.
\item We summarize the primary technical approaches that LLMs offer to the realm of finance, examine the potential in the investment field, and provide a foundational survey for researchers in this domain. 
\item We assess the effectiveness of GPT-4 in various tasks. 
\item We concisely overview of the most significant results from our research, discuss the major unresolved issues that should be addressed in subsequent efforts, and offer insights into future directions and possibilities in this field..
\end{itemize}

\section{Related Work}

\begin{figure*}[t]
\centering
\includegraphics[width=1\textwidth]{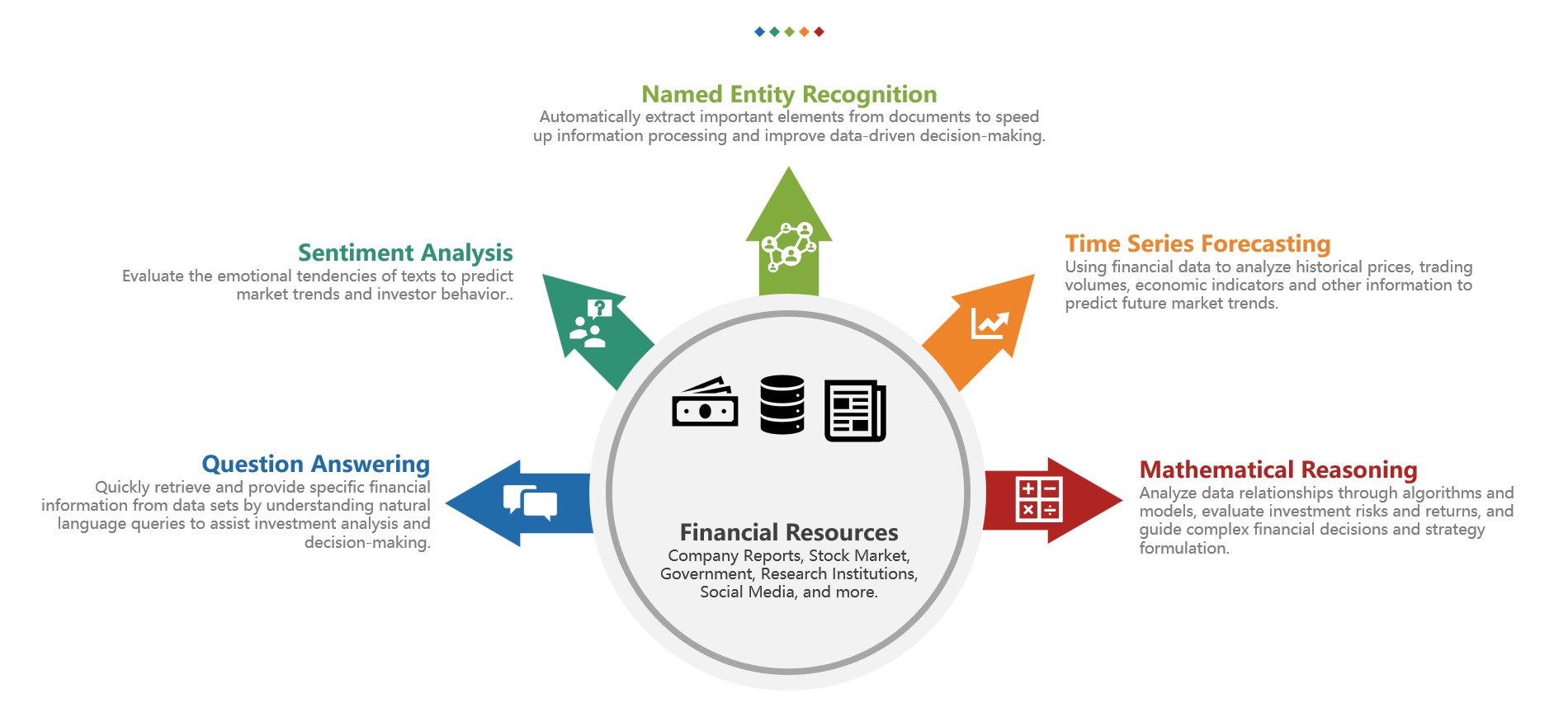}
\caption{The Ability of LLMs in Financial Tasks} 
\end{figure*}

\subsection{Large Language Models}
LLMs are primarily built upon the Transformer architecture  \cite{vaswani2017attention}, which has been central to their ability to handle complex language tasks. The Transformer model is structured with two key components: the Encoder and the Decoder, each consisting of multiple layers of self-attention and feed-forward neural networks. This architecture facilitates effective management of long-range dependencies within sequences.

\begin{equation}
Attention(q, k, v)=softmax(\frac{qk}{\sqrt{d_k}})v
\end{equation}

Self-attention is characterized by its use of queries (Q), keys (K), and values (V), three vectors derived from the input data. Each element in the input sequence is transformed into these three vectors through linear transformation. The self-attention mechanism then computes the attention scores by taking the dot product of the query with all keys. These scores determine how much focus or 'attention' each element in the sequence should have in relation to every other element. The attention scores are normalized using a softmax function, ensuring they sum up to one, thus forming a probability distribution. The final output of the self-attention layer is a weighted sum of the value vectors, where the weights are the softmax-normalized attention scores. This process allows each output element of the self-attention layer to be a combination of the inputs, with the weights specifying the amount of attention given to each input element. The self-attention mechanism's ability to weigh inputs differently allows LLMs to capture complex relationships in the data, such as long-range dependencies, making it exceptionally powerful for tasks that require an understanding of context and sequence.

The architecture of LLMs typically falls into one of two categories: Decoder-only and Encoder-Decoder. Decoder-only models, such as those in the GPT series  \cite{radford2018improving,radford2019language}, generate text in a unidirectional manner  \cite{brown2020language}. Each token in the input sequence attends only to preceding tokens, making them well-suited for tasks like text generation. The Encoder-Decoder models, like T5  \cite{raffel2020exploring} and BART  \cite{lewis2019bart}, feature separate mechanisms for encoding input sequences and decoding them into target sequences. This design allows them to handle a broader range of tasks, including both generation and comprehension.

Token generation in LLMs is a vital process, involving vocabulary creation, probability prediction, and techniques like beam search for sequence generation. Vocabulary in LLMs is typically constructed using methods like Byte-Pair Encoding (BPE)  \cite{gage1994new}, which allows the model to break down words into subword units. This method aids in managing the model's vocabulary size, ensuring efficient handling of rare words and morphemes.

During the token generation process, LLMs predict the probability of each token given the context provided by the input sequence. This is typically achieved through a softmax layer that converts the output logits into a probability distribution over the vocabulary. The model selects tokens based on these probabilities, either choosing the most likely next token (greedy decoding) or leveraging techniques like beam search. Beam search is a decoding strategy  \cite{freitag2017beam} that maintains a fixed number of candidate sequences at each step. It expands each candidate by one token at a time, computes the probability of each expansion, and keeps only the most likely sequences. This method balances between finding the most probable sequence and maintaining a diverse set of candidate sequences, leading to more coherent and contextually appropriate outputs.

The capabilities inherent in the Transformer architecture and token generation processes of LLMs have facilitated their application across a wide range of domains. For instance, in Natural Language Generation (NLG), the Decoder-only models excel at producing contextually relevant text, suitable for creative writing and automated report generation. Encoder-Decoder models, due to their bidirectional processing ability, are highly effective in tasks like machine translation, capable of converting input from one language to another while preserving semantic integrity  \cite{zhao2023survey}.

For example, in conversational AI, LLMs power sophisticated chatbots and virtual assistants \cite{liu2023summary,lee2023multimodality}, capable of generating human-like responses in real-time. Their ability to understand and generate language fluently makes them ideal for customer service automation, interactive learning platforms, and personalized communication tools.

LLMs also play a crucial role in information extraction and summarization  \cite{liu2024understanding,shi2023mededit}, distilling lengthy documents into concise, informative summaries. This application is particularly beneficial in fields like journalism and academic research, where quick assimilation of information is essential.

Furthermore, the sophisticated understanding of context and language nuances allows LLMs to perform sentiment analysis \cite{liu2023summary,liu2024understanding}, including financial sentiment analysis  \cite{zhang2023enhancing}. This capability is widely used in brand monitoring, market research, and social media analysis, providing insights into public opinion and consumer behavior.

Overall, the technical intricacies of LLMs, from their architectural design to their token generation methods, underpin a broad spectrum of applications \cite{liu2023artificial,gong2023evaluating,dai2023auggpt,liu2023deidgpt,holmes2023benchmarking}. These models not only enhance existing processes but also open up new possibilities in the way we interact with and process language.

\subsection{Named Entity Recognition}
Named Entity Recognition (NER) is a key technology in the field of NLP, used to identify and classify entities with specific meanings from text, such as names, places, organizations, time expressions, financial terms, etc. NER plays an important role in information extraction, question-answering systems, content analysis, knowledge graph construction, and other fields  \cite{nasar2021named}.
There are three main mainstream approaches to solving the NER, namely Rule-Based methods, Machine Learning-Based methods, and Deep Learning-Based methods.  \cite{ehrmann2023named} Rule-based systems operate based on identifying entities using predefined rules and patterns, such as using a dictionary of place names to recognize locations. It is easily interpretable and does not require training data. While reliant on expert knowledge, these methods have limited flexibility and scalability  \cite{li2020survey}.
Machine Learning-Based Methods: These methods, such as Support Vector Machines (SVM) and Random Forests, learn to recognize entities through training datasets based on manually selected features. They offer more flexibility than rule-based methods but require extensive annotated data  \cite{ekbal2010named}.
Deep learning techniques for tagging sequences make use of word and character representations that are distributed, by training on sentence or sequence features in an end-to-end manner. These methods mainly use BiLSTM structures or networks based on self-attention. They frequently use a Conditional Random Field (CRF) layer for decoding tags, aiding in the comprehension of label interdependencies. Leveraging these capabilities, deep learning approaches are highly effective in managing intricate patterns and extensive data sets \cite{ehrmann2021named,luo2018attention}.
NER is widely used in the financial field, it can be applied for information extraction (extracting key details about companies, stocks, and market events from financial news and reports), compliance monitoring (automatically identifying and overseeing sensitive entities in financial documents, like money laundering and fraud), and investment decision support (providing data support for investment decisions by analyzing entities and events in market news and reports). These applications underscore the vital role of NER in enhancing efficiency, ensuring compliance, and supporting strategic decisions  \cite{zhao2021bert}.

\subsection{Sentiment Analysis}
In contemporary financial market forecasting, especially regarding Bitcoin trading, the significance of sentiment analysis has been corroborated through numerous academic studies~ \cite{bollen2011twitter,li2014effect,yu2019information}. This research area primarily bifurcates into two methodological categories: lexicon-based and machine-learning approaches, both pivotal in discerning market trends.

\textbf{Lexicon-Based Methodology:} Within this category, approaches are subdivided into dictionary and corpus-based strategies. A notable instance is the model developed by Dev Shah et al.~ \cite{shah2018predicting}, which utilizes the 'pattern' Python library for transforming textual data into numerical vectors. This process involves compiling sentiment scores by quantifying the occurrence of positive and negative words. However, this model faces limitations due to its unweighted sentiment scoring for individual words, potentially leading to inaccuracies in mirroring the actual market sentiment.

\textbf{Machine Learning Techniques:} These are split into unsupervised and supervised learning. The unsupervised model by M.S. Usha et al.~ \cite{usha2013analysis}, which leverages the Gibbs sampling algorithm, excels in identifying sentiment and topics simultaneously. Yet, its inefficacy in capturing neutral sentiments poses a constraint. In contrast, the supervised approach by D.K. Kirange et al.~ \cite{kirange2016sentiment} focuses on classifying emotions in news content to determine sentiment polarity, employing algorithms such as Naive Bayes, SVM, and KNN, with the latter showing optimal accuracy. Moreover, Sneh Kalra et al.~ \cite{kalra2019efficacy} introduced a model that synergizes Naive Bayes sentiment analysis with adjacent date stock variance data from Yahoo Finance, although it is somewhat limited by its reliance on a single data source.
Xiadong Li et al. \cite{li2020incorporating} proposed a novel deep learning-based stock prediction system that fuses sentiment analysis with technical stock indicators. Additionally, the field has seen diverse methodologies such as specialized NLP sub-module designs for sentiment analysis~ \cite{sert2020analysis, nguyen2015sentiment}, the application of N-gram and Naive Bayes Algorithms~ \cite{khedr2017predicting}, dictionary-based sentiment analysis~ \cite{kalyanaraman2014sentiment}, and mood classification paired with daily sentiment scoring~ \cite{mittal2012stock, audrino2020impact}. Time series analysis models have also found their application in this area~ \cite{mohan2019stock}.

These varied methodologies underscore the complexity and multidimensionality of sentiment analysis in financial forecasting, particularly in the context of news analysis. Each approach offers a unique lens through which market trends can be decoded and anticipated, demonstrating the intricate interplay between market sentiment and financial news analysis.

\subsection{Question Answering}

Large language models (LLMs), such as GPT-4, have demonstrated remarkable capabilities in question answering \cite{wei2022chain}, mainly due to their complex architecture and large amounts of training data.

LLMs obtains broad knowledge coverage by analyzing large amounts of text data on the Internet. They can answer questions ranging from general knowledge to specialized fields such as finance, history, science, technology, art, and more \cite{tan2023promises,liu2023transformation,ma2023impressiongpt,liao2023differentiate,dai2023adautogpt,liu2023summary,guan2023cohortgpt,cai2022coarse,liu2023pharmacygpt,shi2023mededit,gong2023evaluating,liu2022survey,cai2023coarsetofine,liao2023maskguided,rezayi2023exploring,liu2023radiology}. LLMs can understand complex queries \cite{wu2023exploring,rezayi2022agribert,liu2023radiologygpt,wang2023review,li2023artificial,holmes2023benchmarking,liu2023transformation,tang2023policygpt,rezayi2022clinicalradiobert}. Whether it’s long sentences, ambiguous questions, or questions that require the synthesis of different information sources, LLMs can handle it and provide relevant answers \cite{liu2023summary,holmes2023evaluating,tang2023policygpt}. LLM can maintain contextual coherence in conversations. This means it can understand and answer subsequent questions based on previous conversations, providing more accurate and relevant information \cite{liu2023summary,liu2023radiologyllama2,zhao2023brain}. Top LLMs often have multilingual capabilities and can understand and answer questions in different languages \cite {tang2023science}, which allows them to serve a wider user base.

LLMs have exhibited remarkable capabilities in advanced reasoning. For instance, GPT-4 showcases its ability for common-sense reasoning by leveraging in-context learning. Moreover, the study \cite{wei2022chain} reveals that when LLMs are provided with well-structured sequential prompts that break down complex, multi-step problems, their performance in tasks involving arithmetic, deductive reasoning, and common-sense understanding improves significantly.

\subsection{Time Series Forecasting}
Financial time series forecasting has traditionally hinged on statistical and econometric methods. Models like ARMA-GARCH have been pivotal in discerning patterns and volatility in financial series \cite{radha2006forecasting}.Over time, these models have been refined to better interpret the intricacies of financial markets. Other methods that have gained prominence include Vector Autoregressive Models (VAM) \cite{zivot2006vector}, State-Space Models utilizing Kalman Filters \cite{orderud2005comparison}, Diffusion Models \cite{fan2005selective}, and Vector Error Correction Model (VECM) \cite{johansen1995likelihood}, forming the bedrock of financial analysis.

The emergence of machine learning has introduced a plethora of models for financial forecasting. Decision trees and support vector machines, known for their effectiveness in financial series prediction, have become particularly prominent. Of late, there has been a pivot towards deep learning techniques such as Recurrent Neural Networks (RNNs), Convolutional Neural Networks (CNNs), and Transformer models, renowned for their proficiency in unraveling complex, non-linear data relationships.

The development of LLMs like GPT-3 \cite{brown2020language}, GPT-4 \cite{achiam2023gpt}, and LLaMA \cite{touvron2023llama}, has been a game-changer in the realm of financial time series forecasting. These models excel in parsing and interpreting intricate dependencies in diverse data sets, offering outputs that are comprehensible to humans. There has been considerable advancement in this domain, including the conversion of time series data into textual sequences, the creation of varied prompts for intelligible financial forecasting, and the conceptualization of financial time series as multimodal data, harnessing the combined strengths of LLMs and computer vision. These developments showcase the dynamic and expanding role of LLMs in financial time series forecasting, highlighting a field ripe with innovation and exploration \cite{yu2023temporal,jin2023time,chang2023llm4ts}.


\subsection{Mathematical Reasoning}

Mathematical reasoning forms the cornerstone of modern finance, serving as the bedrock upon which complex financial theories, models, and practices are constructed. In the realm of finance, mathematical reasoning extends beyond mere number crunching; it encompasses the application of mathematical principles to analyze and solve financial problems, thereby empowering professionals to make informed decisions, assess risks, and forecast market trends.

Central to mathematical reasoning in finance is the integration and application of various mathematical disciplines, such as calculus, statistics, probability, and linear algebra. These mathematical frameworks enable finance professionals to devise and interpret financial models, assess investment strategies, and optimize portfolios. Calculus, for instance, is pivotal in modeling the dynamic behavior of markets and in calculating derivatives, which are key in risk management and the pricing of financial instruments \cite{brigo2006interest}. Moreover, statistics and probability are indispensable in evaluating risks and returns, aiding in asset valuation and the development of predictive models \cite{lee2019financial}.

Furthermore, mathematical reasoning in finance is dynamic and continually evolves with the emergence of new theories and the advent of technological advancements. The inception of quantitative finance, which amalgamates mathematical finance, numerical methods, and computer simulations, has transformed the industry. This interdisciplinary approach has led to the creation of intricate models for options pricing, risk management, and algorithmic trading, thereby enhancing the precision and efficiency of financial operations \cite{buehler2019deep}.

As we traverse an era marked by increasing complexity and interconnectivity in financial markets, the significance of mathematical reasoning becomes increasingly critical. It not only furnishes finance professionals with the necessary tools for understanding and innovation but also instills a rigorous analytical framework, which is vital amidst financial uncertainties. Whether it's in the valuation of complex derivatives, the formulation of robust financial models, or the strategic management of investment portfolios, mathematical reasoning remains an essential component in the repertoire of contemporary finance \cite{cont2001empirical}.

\section{Scope of Finance Tasks}
\subsection{Financial Engineering}
Financial Engineering is a multidisciplinary field that combines finance, mathematics, and computer science to create and implement innovative financial strategies and products. LLMs assist in Financial Engineering by enhancing two key subtasks: Quantitative Trading and Portfolio Optimization. 
\subsubsection{Quantitative Trading}
Quantitative trading has traditionally relied on mathematical and statistical models to drive investment decisions, often centering around historical data and predefined algorithmic strategies. This approach, while effective in certain market conditions, faces challenges in dynamic and complex market environments. Traditional quantitative models can struggle to adapt quickly to new information, particularly when it comes to unstructured data sources like news articles, social media, and financial reports. These sources contain valuable sentiment and opinion-based information that standard quantitative methods may overlook \cite{liu2023fingpt,yu2023finme}.

In recent years, the emergence of LLMs has opened new avenues in quantitative trading. LLMs, with their advanced natural language processing capabilities, play a pivotal role in effectively extracting and utilizing such implicit sentiment information in investment strategies. By analyzing vast amounts of textual data, LLMs can identify subtle, often nuanced sentiments embedded in analysts' reports, market news, and financial statements. These sentiments are crucial as they often represent the collective market sentiment and can precede major market movements.

Analysts' reports, for instance, are a goldmine of insights but are often laden with implicit sentiments that the analysts might be reluctant or avoid revealing directly. LLMs can decipher these subtle cues, providing a more comprehensive understanding of market dynamics. This capability extends beyond mere sentiment analysis; it encompasses the understanding of context, the detection of sarcasm, and the interpretation of complex financial jargon, which are often lost in traditional quantitative analysis.

The integration of LLMs into quantitative trading strategies represents a significant advancement in the field. It allows for a more holistic approach to investment decisions, one that combines the precision of quantitative models with the nuanced understanding of market sentiments. This synergy not only enhances the robustness of trading strategies but also provides a competitive edge in rapidly changing market conditions. As the financial markets continue to evolve, the role of LLMs in quantitative trading is poised to become increasingly vital, marking a paradigm shift in how investment decisions are made.
 \cite{wang2023alphagpt}
    
\subsubsection{Portfolio Optimization}
Traditional portfolio optimization, grounded in the principles of modern portfolio theory, seeks to balance risk against return, typically relying on historical market data and statistical analysis \cite{paulson1991makes}. This approach, while systematic, often encounters challenges in rapidly evolving markets where historical data may not adequately predict future trends. Additionally, traditional models may not fully account for complex, real-world factors like geopolitical events or sudden market shifts, potentially leading to suboptimal asset allocations \cite{aligeopolitical}.

The integration of LLMs in portfolio optimization heralds a significant advancement in addressing these challenges. LLMs excel in processing and analyzing vast amounts of unstructured data, including market reports, news articles, and financial statements, providing deeper insights and supplementary analysis crucial for risk assessment. These models can uncover subtle market sentiments and emerging trends hidden in textual data, offering a more nuanced view of potential risks and opportunities. By augmenting quantitative data with qualitative insights derived from LLMs, investors can achieve a more holistic approach to portfolio optimization. This synergy not only enhances the robustness of traditional models but also equips investors with a more adaptive and informed strategy in the face of market uncertainties. \cite{jeong2024finetuning}
\subsubsection{Robo-advisors}
Leveraging the analytical power of LLMs and artificial intelligence (AI), robo-advisors are making significant strides in reshaping the world of financial investing. Combining precision, adaptability, and accessibility, these advanced platforms are quickly becoming popular tools for wealth management and investment advisory services.

The essence of robo-advisors’ appeal lies in their computational power, which allows them to tailor portfolios to the individual user’s circumstances, taking into account market dynamics and personal risk preferences. The LLMs is critical in this context, parsing extensive data sets to discern complex financial market patterns, allowing robo-advisors to provide informed investment guidance. Throughout the investment cycle, they continuously monitor portfolio performance, adjusting the balance between expected returns and user-defined risk thresholds \cite{caspigenerative}.

A key benefit of robo-advisors is their ability to flexibly update investment strategies to reflect changes in the market, a flexibility often not available with traditional investing avenues. The enhanced flexibility can foster greater trust between financial advisors and their clients \cite{baker2017regulating}.While historical analysis has primarily focused on the algorithmic sophistication and legitimacy of robo-advisors  \cite{lourencco2020whose}, discussion is burgeoning about the psychological factors that guide individuals to use these AI platforms. 

In a revealing study of the German robo-advisory market (covering approximately 78 assets), they examined approximately 243,000 portfolio pairs along with customer demographic data. The findings indicate that despite the high level of AI-driven sophistication, the personalization aspects of robo-advisory advice are currently limited \cite{scherer2023trust}. Key factors that influence modern portfolio choices—such as the amount and nature (beta) of human capital or shadow assets—remain largely unresolved. Recommendations tend to cater to current investor biases or regulators’ views on portfolio allocation, which inadvertently limits the economic potential of robo-advisors while bolstering consumer confidence and ensuring regulatory sanctions. The renaissance of robo-advisory advice is highlighted by its tendency to eschew complex, customized strategies in favor of more broadly applicable investment principles, for reasons including explainability to the average user and the need for privacy and data security \cite{aw2023counteracting}.

All in all, the integration of LLMs with robo-advisory services marks a quantum leap in the field of consulting. These AI-centric platforms will revolutionize investing and wealth management by connecting intricate financial acumen with the understanding of ordinary investors, although there is caution about their current scope for customization \cite{maple2023ai}  \cite{nourallah2023one}.

\subsection{Financial Forecasting}

\subsubsection{Merge and Acquisition Forecasting}

In Mergers and Acquisitions (M\&A) forecasting, NLP offers pivotal tools for mining and interpreting vast arrays of textual data  \cite{yang2021fact,visintin2023leveraging}. LLMs can adeptly analyze financial reports, news articles, and press releases to unearth underlying trends or strategic shifts that may hint at forthcoming M\&A activities  \cite{yang2023fingpt}. 

Furthermore, sentiment analysis, a crucial facet of NLP, scrutinizes market commentaries and financial reports. This analysis is instrumental in detecting shifts in market sentiment regarding specific companies or sectors, potentially foreshadowing M\&A endeavors. 

Additionally, LLMs can delve into historical M\&A cases and identify linguistic and financial patterns that typically precede such corporate actions. This historical insight is invaluable in predicting future M\&A activities. Lastly, the role of social media cannot be understated. LLMs can monitor these platforms for speculative information and public sentiment, often serving as early indicators of possible M\&A movements.

\begin{itemize}
    \item \textbf{Analyzing Financial Reports and News Articles:} Hypothetical Scenario: LLMs analyze the financial reports and news articles surrounding tech giants like Apple (AAPL) and a smaller, innovative tech company like Roku (ROKU). The analysis reveals a pattern of increasing mentions of collaborative projects and shared technology initiatives, suggesting a strategic alignment. This could hint at a potential acquisition of Roku by Apple, a move that could significantly expand Apple's footprint in the streaming hardware market.

    \item \textbf{Sentiment Analysis of Market Commentaries:} Hypothetical Scenario: NLP tools conduct sentiment analysis on market commentaries regarding the pharmaceutical industry. They detect a positive shift in sentiment towards Merck (MRK) and a smaller biotech firm, BioNTech (BNTX), known for its breakthroughs in mRNA technology. This sentiment shift, coupled with increased collaborative research efforts between the two, might suggest an impending merger or partnership, aligning Merck's robust distribution network with BioNTech's innovative vaccine technology.

    \item \textbf{Examining Historical M\&A Patterns:} Hypothetical Scenario: An LLM reviews historical M\&A cases in the automotive sector, particularly focusing on Tesla (TSLA) and its past acquisitions of smaller tech companies specializing in autonomous driving technology. By identifying linguistic and financial patterns from these cases, such as Tesla's strategic investments in AI technology, the model predicts Tesla’s interest in acquiring a company like Mobileye (MBLY), a leader in advanced driver-assistance systems.

    \item \textbf{Monitoring Social Media for Speculative Information:} Hypothetical Scenario: LLMs monitor platforms like Twitter and LinkedIn for discussions involving Disney (DIS) and Netflix (NFLX). An uptick in speculative discussions about Disney's interest in enhancing its streaming content and a potential strategic fit with Netflix's vast content. 
    
\end{itemize}

\subsubsection{Insolvency Forecasting}

For insolvency forecasting, language models can analyze a myriad of textual sources to gauge a company's financial health accurately. By evaluating financial disclosures, news articles, and statements from corporate leaders, these models can detect early signs of financial distress  \cite{chen2018models}. 

To complement traditional numerical modeling, these models integrate textual analysis from various reports and news sources into bankruptcy prediction models, enhancing their predictive accuracy  \cite{ahmadi2018towards}. The sentiment and tone in corporate communications and financial discussions can be meticulously analyzed and reveal early warnings of a company's deteriorating financial situation. 

Moreover, a critical examination of regulatory filings through NLP can reveal subtle linguistic or disclosure patterns  \cite{kim2021corporate}. These patterns are frequently observed in the prelude to financial difficulties or impending insolvency, providing essential insights for stakeholders and investors.

\begin{itemize}
    \item \textbf{Analyzing Financial Health of Retail Companies:} LLMs can assess the financial statements of retail companies. They would identify signs of financial distress, such as declining sales and increasing debt levels, that may indicate a risk of insolvency.

    \item \textbf{Sentiment Analysis in Industry News:} Sentiment analysis on news articles and financial reports about a technology firm. If there's a prevalent negative sentiment and discussions about liquidity issues or declining market share, this could signal financial troubles ahead.

    \item \textbf{Monitoring Social Media for Consumer Sentiment:} LLMs can track social media mentions of an automotive company, say, "AutoDrive Motors". By analyzing consumer sentiment and discussions about product issues or declining brand popularity, potential financial struggles could be anticipated.

    \item \textbf{Analyzing Credit Ratings and Analyst Reports:} A language model can examine changes in credit ratings and analyst reports to spot trends such as credit rating downgrades or negative outlooks by financial analysts could be early indicators of impending financial difficulties.

    \item \textbf{Reviewing Legal and Regulatory Filings:} LLMs can analyze legal and regulatory filings for a pharmaceutical company. Increases in litigation cases or regulatory fines might be early signs of financial instability.
    
\end{itemize}

\subsubsection{Market Trend Forecast}
Incorporating GPT-4's capabilities into market trend analysis represents a significant leap forward in the application of artificial intelligence within the domain of financial forecasting. The endeavor to leverage GPT-4's API for predicting stock price trajectories is an intricate process that navigates through a confluence of challenges.

Historically, the academic sphere has gravitated towards econometric models, such as ARIMA  \cite{nelson1998time}, and the finance industry has harnessed machine learning algorithms to predict stock movements. These methods, while effective to a degree, cannot often evolve rapidly with market conditions or explain their predictions transparently.

Market dynamics are notoriously difficult to predict due to their stochastic nature and the multitude of influential variables ranging from macroeconomic indicators to geopolitical events and investor sentiment. These factors are interdependent and can exhibit non-linear relationships, presenting a daunting task for any predictive model. Traditional quantitative models, while robust, often struggle to account for the subtleties of market sentiment and the rapid shifts in global economic landscapes.

NLP is increasingly being utilized in market forecasting to complement traditional quantitative analysis methods  \cite{lopez2023can}. By processing and interpreting textual data from various sources such as news articles, financial reports, and social media, NLP provides valuable insights into market sentiment and trends  \cite{yang2023fingpt}. This use of NLP helps in uncovering underlying patterns and correlations in market behavior that might not be immediately apparent from numerical data alone. In market forecasting, the ability of NLP to quickly analyze large volumes of text and extract relevant information plays a crucial role in making timely and informed predictions about market movements.
\begin{figure*}[t]
\centering
\includegraphics[width=1\textwidth]{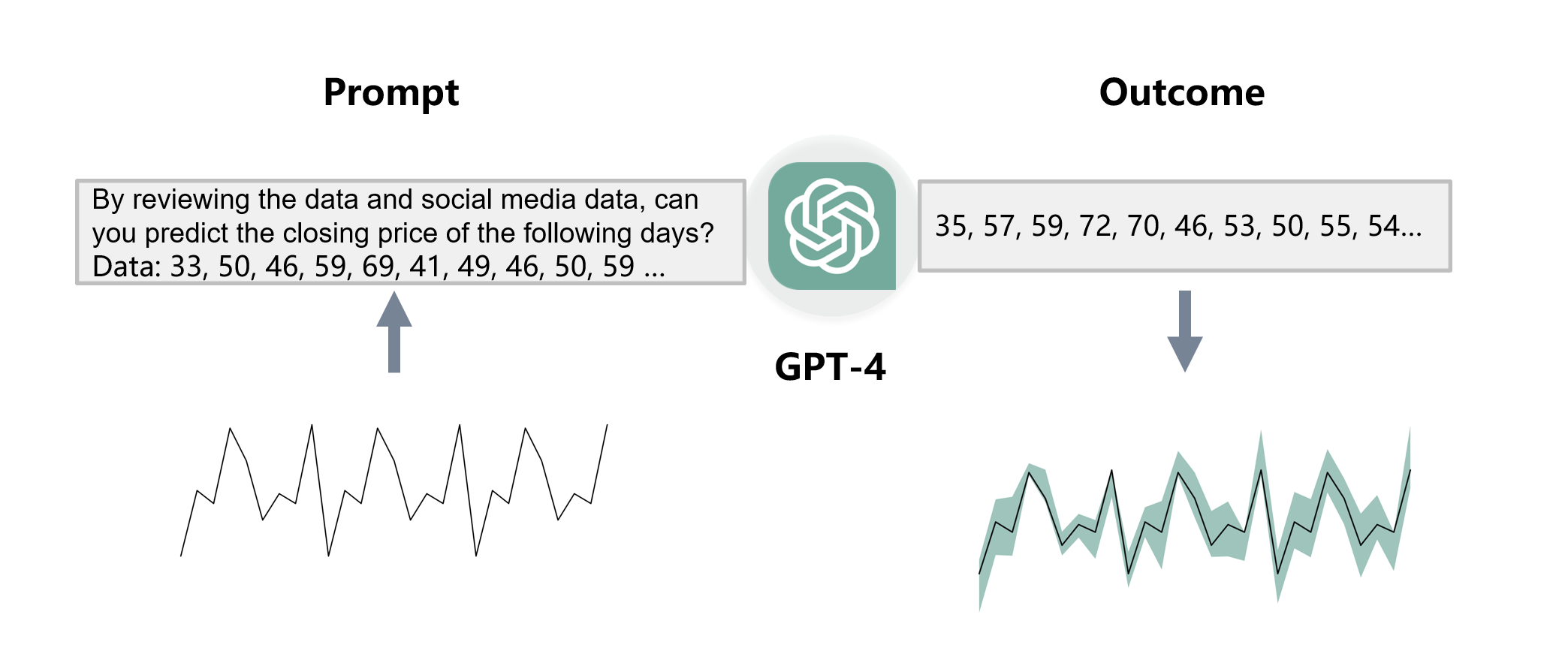}
\caption{GPT-4's forecasting capability on stock price movement} 
\end{figure*}

Diverse Data Sources in Market Forecasting with LLMs:
\begin{itemize}
  \item \textbf{Processing Financial News and Reports:} LLMs can quickly digest and analyze extensive financial news and reports, providing a comprehensive view of market conditions and potential trends.
  \item \textbf{Sentiment Analysis of Social Media:} By examining social media platforms and analyzing the sentiment of posts and tweets, LLMs can gauge public opinion and investor sentiment, which are crucial indicators of market movements.
  \item \textbf{Interpreting Economic Indicators:} LLMs can interpret textual data related to economic indicators such as inflation rates, employment data, and GDP growth, which traditionally influence market forecasts.
  \item \textbf{Scenario Simulation:} Leveraging historical data, LLMs can simulate various market conditions and outcomes, aiding in risk assessment and decision-making processes.
  \item \textbf{Real-time Data Processing:} The ability to process data in real time allows LLMs to stay abreast of rapid market changes, offering timely insights for forecasting.
\end{itemize}

Advantages and Potential of LLMs in Market Analysis:
\begin{itemize}
  \item \textbf{Enhanced Predictive Capabilities:} By analyzing a broader range of data sources, LLMs can offer more accurate predictions than traditional numerical-only methods.
  \item \textbf{Holistic Market Understanding:} The integration of textual data analysis provides a more holistic understanding of market dynamics, beyond what numerical data alone can offer.
  \item \textbf{Adaptability to Market Changes:} The AI-driven nature of LLMs allows for quick adaptation to new information and changing market scenarios.
  \item \textbf{Customizable Analysis:} LLMs can be tailored to focus on specific sectors, regions, or types of data, making them versatile tools for various market analysis needs.
  \item \textbf{Reducing Human Bias:} By relying on data-driven insights  \cite{makridakis2023large}, LLMs can help reduce human bias in market forecasting, leading to more objective and reliable predictions.
\end{itemize}

As the most powerful language model to date  \cite{sun2024trustllm,wang2024large}, GPT-4 brings to the table its formidable prowess in processing vast datasets, extracting nuanced patterns, and synthesizing this information to generate predictions. Its capacity to parse through disparate data sources, including real-time financial news, historical price data, and burgeoning trends on social media platforms, allows it to construct a multi-faceted view of market conditions.

Furthermore, GPT-4 transcends mere predictive output; it provides the underlying rationale for its forecasts, thereby granting investors and analysts a window into the 'thought process' of the AI. This interpretability is paramount, as it aligns with the rigorous standards of academic research and financial scrutiny, enabling stakeholders to make informed decisions.

The experiment conducted with GPT-4's API, which culminated in accurate and interpretable outcomes, indicates a paradigm shift.  This advancement may redefine predictive analytics in finance, offering a more dynamic, holistic, and transparent approach to understanding and anticipating market trends. This study serves as a testament to the potential of integrating advanced AI into financial analysis and the broader implications for future research and practical applications within the industry.

\subsection{Financial Risk Management}
\subsubsection{Credit Scoring}

The significance of credit and risk assessments in the financial sector cannot be overstated, as these evaluations play a crucial role in maintaining financial stability. Credit assessment not only covers the possibility of assessing an individual borrower's ability to repay, but also includes a variety of applications such as analyzing the risks of potential investments, evaluating the financial health of a company, and assisting financial institutions in making decisions about loan policies and interest rates. Traditionally, financial credit and risk assessment methods have predominantly been rule-based or reliant on machine learning algorithms \cite{dastile2020statistical}. However, these approaches exhibit limited flexibility across different tasks, often being tailor-made for specific objectives. Consequently, they struggle to generalize or integrate knowledge from diverse financial tasks. Moreover, such methods cannot leverage insights transferable across various financial activities. The advent of LLMs offers a promising avenue to transcend these limitations \cite{yoon2023design,babaei2023gpt,zheng2020effects}. LLMs, with their prowess in multi-task learning and few-shot generalization, present an opportunity to redefine the landscape of financial assessments. Current research is exploring the potential of LLMs to identify correlations between disparate financial tasks and generalize across them. This capability marks a potential paradigm shift in credit and risk evaluation methodologies. The application of LLMs in this domain, however, is not without its challenges. For instance, the need to analyze tabular data, which contains symbolic information markedly different from the natural language data that LLMs are typically trained on, presents a significant hurdle. Additionally, ensuring that these models avoid biases in sensitive attributes such as age or gender is paramount. Despite these challenges, LLMs offer considerable advantages in processing and analyzing large volumes of textual data, such as loan applications and transaction histories. This capability enables them to extract valuable insights that can be instrumental in credit and risk analysis. By analyzing historical data and market trends, LLMs can assist analysts in gaining a deeper understanding of market dynamics and individual credit risks. Nevertheless, it's important to recognize that the effectiveness of LLMs in credit scoring is still evolving. As the financial industry continues to integrate more advanced technological solutions, the role of LLMs in enhancing the accuracy and efficiency of credit and risk assessments will likely become more pronounced, heralding a new era in financial analytics.
\subsubsection{ESG Scoring}
Environmental, Social, and Governance (ESG) scoring is a critical metric in the contemporary business and investment landscape. It serves as a tool for evaluating a company's commitment to environmental stewardship, social responsibility, and governance practices. ESG scores came about due to the financial world’s need to assess companies against these three criteria to identify the best performers in these aspects \cite{clement2022improving}. Private commercial firms whose primary clients are portfolio managers and other investors use tangible and intangible data to construct ESG scores to produce new data that meet investors’ needs \cite{escrig2019rating}.

A growing trend sees more companies being evaluated by sustainability rating agencies. The objective of these assessments is to generate relevant data for stakeholders interested in utilizing non-financial information about these companies. The information is particularly valuable for those looking to assess their investments or to develop investment portfolios based on sustainability criteria \cite{friede2015esg}. There are several common approaches for ESG scoring. Firstly, companies like Refinitiv and Bloomberg collect data from public sources but do not offer any value-adding input or scoring \cite{zumente2021esg}. Secondly, ESG data providers combine both public and own-created data to evaluate ESG scores or ratings (e.g. MSCI). Thirdly, some companies focus on specialized ESG issues such as Carbon Disclosure Project.

The integration of GPT-4 into the process of ESG scoring remains an abundant blank area deserving to explore. Application of GPT-4 can offer numerous potential benefits, enhancing both the efficiency and effectiveness of this increasingly important evaluation method. GPT-4 assists with enhanced data processing and analysis. Its advanced capabilities allow it to process vast amounts of unstructured data rapidly including corporate sustainability reports, news articles, social media posts, and other relevant documents. By analyzing this data, GPT-4 can extract key insights about a company's ESG practices, providing a more comprehensive view than traditional methods. GPT-4 helps mitigate human biases which leads to more objective and consistent ESG assessments. Its ability to analyze data based on predefined criteria reduces subjective interpretation, improving the credibility of the scoring process. Besides, GPT-4 is suitable for real-time monitoring and dynamic scoring. GPT-4 can continuously monitor various data sources for real-time updates related to ESG factors. Dynamic scoring is available to reflect the most current information, providing a more accurate and timely picture of a company's ESG performance.

Utilizing GPT-4 for ESG scoring represents a significant advancement in sustainability evaluation. Its ability makes it a potent tool for providing deeper, more accurate, and up-to-date insights into ESG performance.
\subsubsection{Fraud Detection}
As trade volume escalates and digital wallet technology advances, the realm of financial risk management is increasingly facing sophisticated high-tech criminal activities. A case in point: during the initial nine months of 2023, over 83,000 Americans fell prey to credit card fraud, leading to collective financial losses for the victims amounting to \$183 million \footnote{\url{https://www.aura.com/learn/credit-card-scams}}. Consequently, the implementation of robust fraud detection applications is imperative to preserve the integrity of financial systems and safeguard both the institutions and their clients from financial losses  \cite{feng2023empowering}. Leveraging advanced reasoning and text mining capabilities, LLMs can significantly contribute to the identification of financial fraud in various domains including transactions, emails, profiles, contractors, and decentralized finance  \cite{luo2023ai}. These LLMs serve as an initial filter, learning from customer transaction histories and detailed transaction information to isolate highly suspicious transactions from the billions processed  \cite{roy2018deep}, thereby substantially alleviating the manual labor burden involved in investigating vast quantities of transaction data. In this study, we use the PaySim simulates mobile money transactions dataset  \cite{lopez2016paysim} to evaluate GPT-4's effectiveness in detecting fraud.

\subsubsection{Compliance Check}
LLMs with zero-shot learning capabilities are becoming indispensable in the dynamic world of financial compliance, where regulations are in a constant state of flux. Zero-shot LLMs can adapt to new standards without the need for fine-tuning, which traditionally demands regular updates and a wealth of annotated data \cite{zheng2023llm}. This characteristic is particularly beneficial for tasks such as audits, transaction monitoring, and reporting, as well as financial reporting and disclosure. In audits, zero-shot LLMs can immediately be deployed to parse and analyze documents, identifying inconsistencies and irregularities by understanding the underlying context, without the need for a model that is fine-tuned to specific audit criteria which may change over time \cite{liu2023gpt}. This saves valuable time and resources in an environment where regulatory frameworks can shift unpredictably. For transaction monitoring and reporting, these LLMs excel at detecting anomalous patterns indicative of non-compliance or suspicious activities. They are capable of understanding transactional nuances and alerting to irregularities, all without prior fine-tuning to the specific rules that could be subject to change due to evolving regulations or market practices. When it comes to financial reporting and information disclosure, zero-shot LLMs ensure that disclosures align with current reporting standards such as IFRS, even as those standards are updated. They provide an agile response to changing requirements, highlighting discrepancies against the latest regulations without the need for retraining on new data sets. The zero-shot learning approach of LLMs not only bypasses the labor-intensive process of continuous model retraining but also mitigates the risk of outdated compliance checks in the rapidly changing financial landscape. This agility makes zero-shot LLMs a critical tool for financial institutions seeking to maintain compliance with the latest regulatory demands \cite{clement2022improving,kojima2022large}.

\begin{figure*}[t]
\centering
\includegraphics[width=1\textwidth]{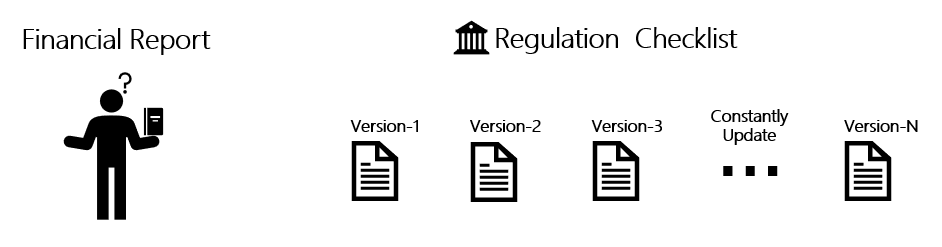}
\caption{In compliance checks, due to rapid updates in regulation checklists, models fine-tuned on outdated standards quickly become obsolete. Therefore, we increasingly rely on GPT-4's zero-shot learning capabilities.} 
\end{figure*}

\subsection{Financial Real-Time Question Answering}

\subsubsection{Financial Education}
GPT-4 is an advanced artificial intelligence language model developed by OpenAI that is capable of understanding and generating human-like natural language \cite{dai2023auggpt,liu2023deidgpt,ma2023impressiongpt,holmes2023evaluating,liao2023differentiate,dai2023adautogpt,liu2023summary,guan2023cohortgpt,cai2022coarse,liu2023pharmacygpt,shi2023mededit,gong2023evaluating,li2023artificial,liu2023context}. This feature makes it a powerful tool for financial education, especially when it comes to explaining complex financial concepts, providing customized learning experiences, and enhancing user interaction \cite{wu2023bloomberggpt}.

First, GPT-4 can simplify complex financial concepts into easy-to-understand language. The field of finance is full of complex terms and concepts such as securities markets, portfolio diversification, risk management, etc. GPT-4 can explain these concepts more understandably through its deep learning and training on large amounts of financial data. This is especially important for those new to finance because it lowers the learning curve, allowing them to more easily understand and apply the concepts.

Secondly, GPT-4 has unique advantages in providing customized learning experience  \cite{10.1007/978-3-031-45673-2_46,liu2023radiologyllama2,tang2023policygpt,liu2023radoncgpt,liuradiology,LIU2023100045,dou2023artificial,holmes2023evaluating}. It can adjust content and difficulty according to the user's learning progress, interests and needs \cite{gong2023evaluating,liu2023transformation,holmes2023benchmarking,shi2023mededit,zhong2023chatabl,zhou2023fine,liao2023mask,zhao2023ophtha}. For example, for beginners, GPT-4 can provide basic financial knowledge and concepts; for more experienced learners, it can provide more in-depth analysis and advanced topics. This personalized learning approach helps improve learning efficiency and user satisfaction.

In addition, GPT-4 plays an important role in enhancing user interaction. Through interactive Q\&A, simulated scenarios, and real-time feedback, GPT-4 can create a more dynamic and engaging learning environment. This method not only enhances the interest of learning but also helps improve learners' practical skills and problem-solving abilities.

However, although GPT-4 has many advantages in financial education, it also has some limitations. First, although GPT-4 is excellent at explaining financial concepts and providing personalized teaching, it still relies on existing knowledge bases and data. This means that GPT-4 may not be able to provide cutting-edge information when faced with the latest financial trends and data. For example, in the context of rapidly changing financial markets, GPT-4 may not be able to update its knowledge base promptly to reflect the latest market dynamics and regulatory changes. Secondly, GPT-4 also needs to consider ethical and compliance issues when providing financial education. The accuracy and transparency of financial information are critical to protecting consumers and maintaining market order. Therefore, when using GPT-4 as a financial education tool, you must ensure that the information and advice it provides comply with relevant laws and regulations and are ethically responsible.

Overall, GPT-4 offers many potential advantages in the field of financial education, including simplifying complex concepts, providing a personalized learning experience, and enhancing user interaction. However, its application also needs to take into account challenges such as accuracy, timeliness, and ethical compliance. In the future, with the continuous development and improvement of technology, GPT-4 has the potential to become an important auxiliary tool in the field of financial education, helping users to better understand and apply financial knowledge.

\section{GPT-4 Empowered Financial Tasks Evaluations}
In this section, we introduce the approach used in our survey to evaluate the performance of GPT-4 in a variety of financial tasks with one-shot learning and zero-shot prompting. Our method consists of several important parts, including practical financial tasks, selection of benchmark data sets, design of various instruction prompts, and selection of evaluation indicators.
\subsection{Datasets}

To showcase the extensive capabilities of GPT-4 in the financial sector, we have meticulously chosen six diverse datasets. These datasets encompass a wide range of text types, including news articles, analytical reports, and social media posts like tweets. In addition, we've incorporated featuring time series, tabular data, and textual content. Furthermore, we've crafted a series of comprehensive and practical financial tasks that mirror real-world finance scenarios. This selection and design aim to fully demonstrate the advanced and versatile applications of these AI models in finance.
\begin{table}[ht]
\centering
\caption{The details of the raw data and instruction data.}
\resizebox{\textwidth}{!}{%
\begin{tabular}{lcccccc}
\hline
Data & Task & Raw & Instruction & Data Types & Modalities & License \\
\hline
FPB & sentiment analysis & 4,845 & 48,450 & news & text & CC BY-SA 3.0 \\
FiQA-SA & sentiment analysis & 1,173 & 11,730 & news headlines,tweets & text & Public \\
NER & named entity recognition & 1,366 & 13,660 & financial agreements & text & CC BY-SA 3.0 \\
FinQA & question answering & 8,281 & 8,281 & earnings reports & text,table & MIT License \\
ConvFinQA & question answering & 3,892 & 3,892 & earnings reports & text,table & MIT License \\
BigData22 & stock movement prediction & 7,164 & 7,164 & tweets,historical prices & text,time series & Public \\
\hline
\end{tabular}%
}
\end{table}
\begin{figure*}[t]
\centering
\includegraphics[width=1\textwidth]{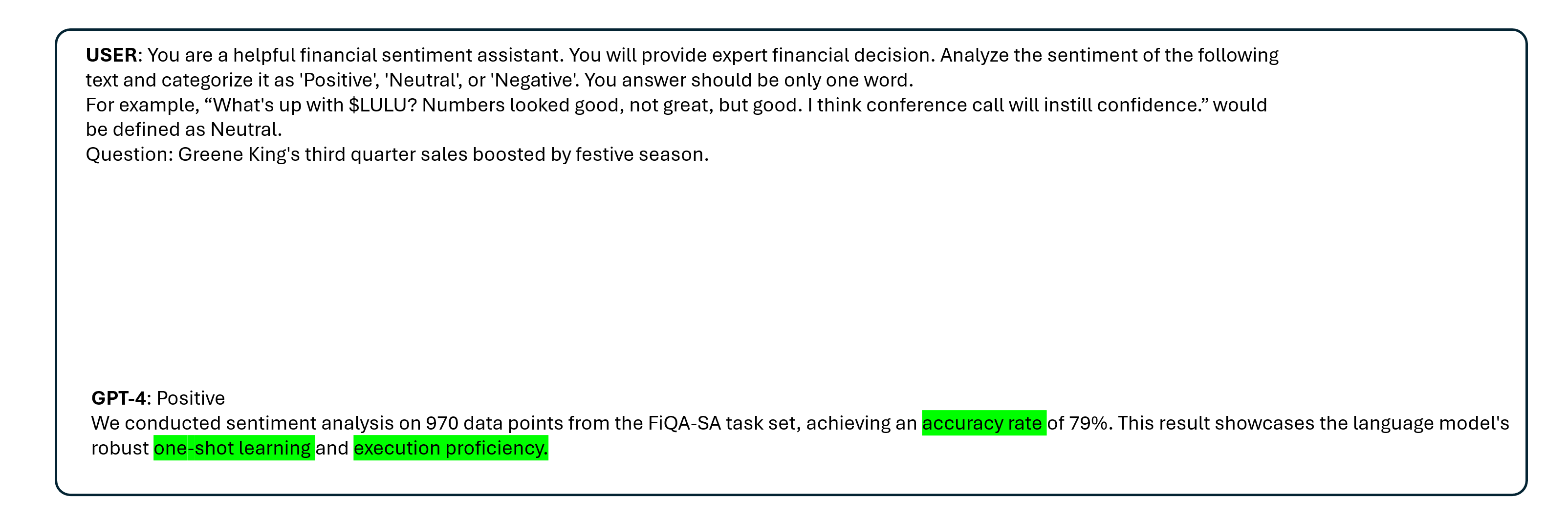}
\caption{We conducted sentiment analysis on 970 data points from the FiQA-SA task set\cite{xie2023pixiu}. By using GPT-4, we achieve 79\$ accuracy rate.} 
\end{figure*}

\begin{figure*}[t]
\centering
\includegraphics[width=1\textwidth]{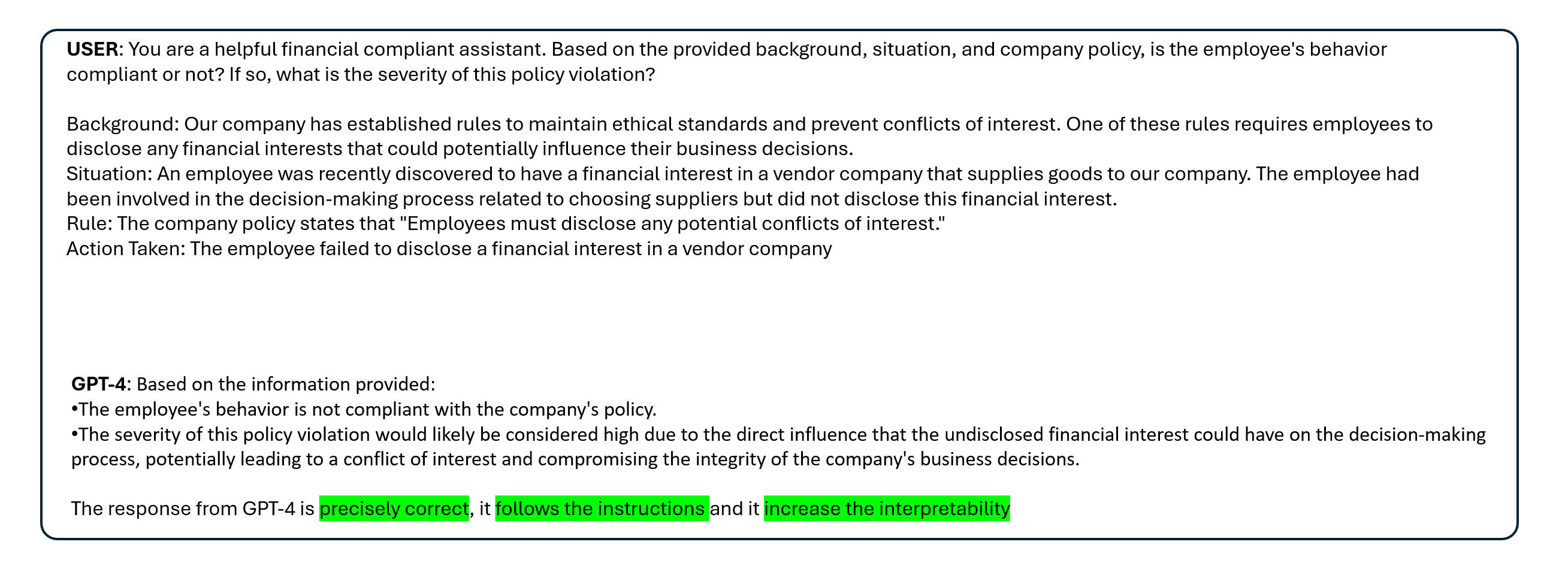}
\caption{GPT-4 has demonstrated its zero-shot learning and instruction following capacities.} 
\end{figure*}

\begin{figure*}[t]
\centering
\includegraphics[width=1\textwidth]{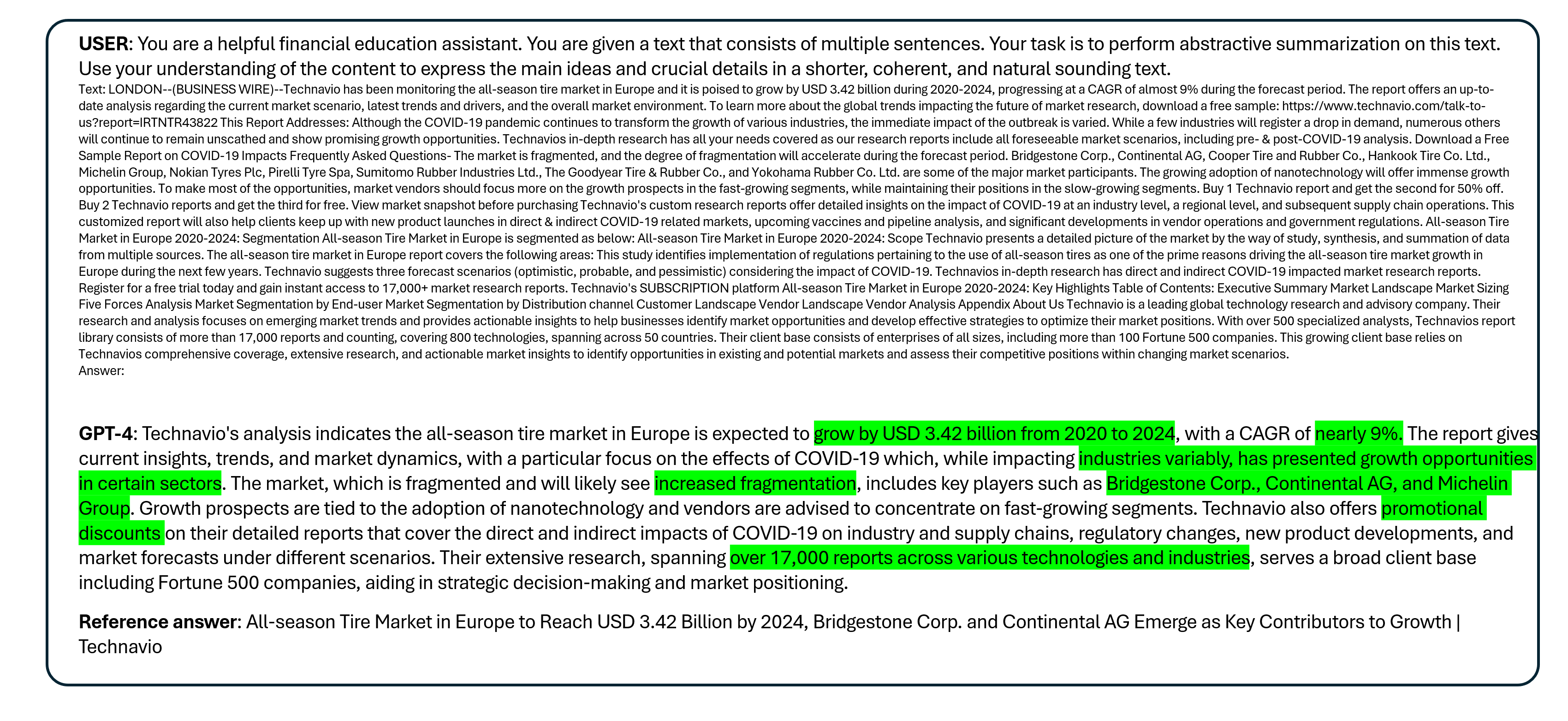}
\caption{Using GPT-4 for summary demonstrates LLMs' knowledge integration capabilities, logical reasoning capabilities and language expression capabilities.Green highlight indicates the k} 
\end{figure*}

\begin{figure*}[t]
\centering
\includegraphics[width=1\textwidth]{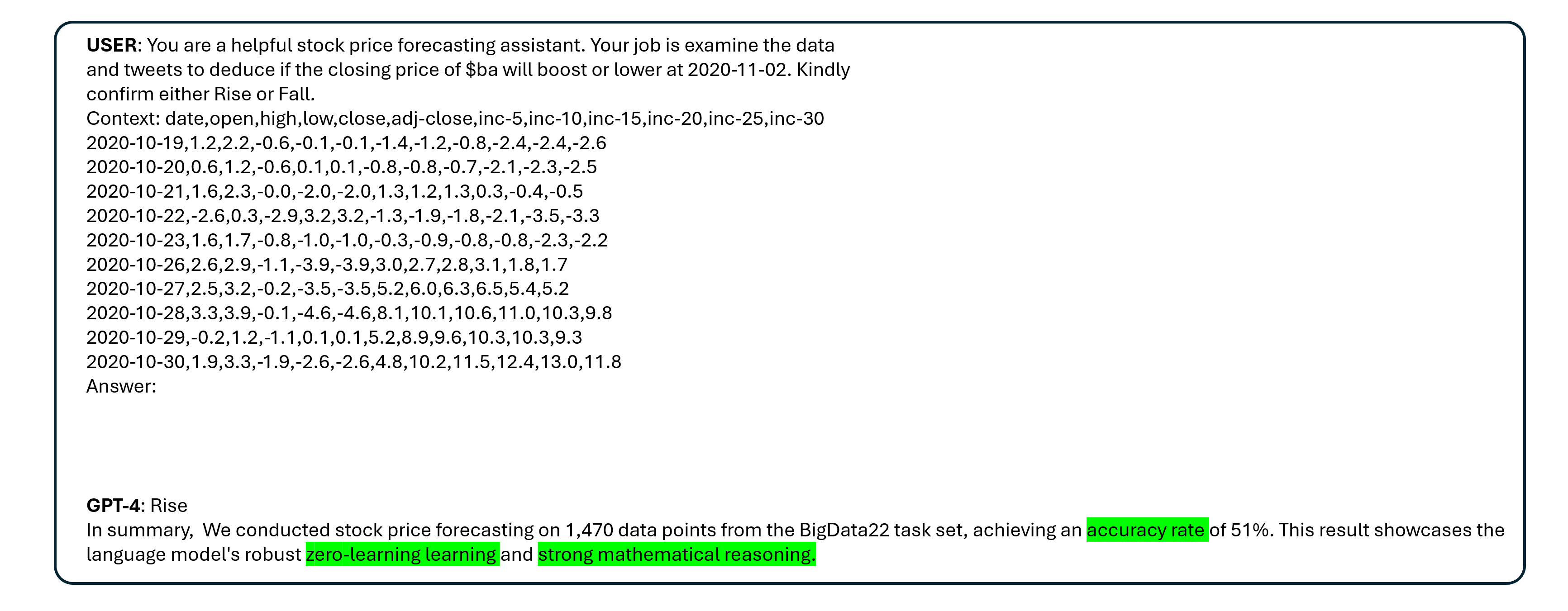}
\caption{We conducted stock price prediction on 1,470 data points from the BigData task set\cite{xie2023pixiu}. By using GPT-4, we achieve 51\$ accuracy rate.} 
\end{figure*}
\begin{figure*}[t]
\centering
\includegraphics[width=1\textwidth]{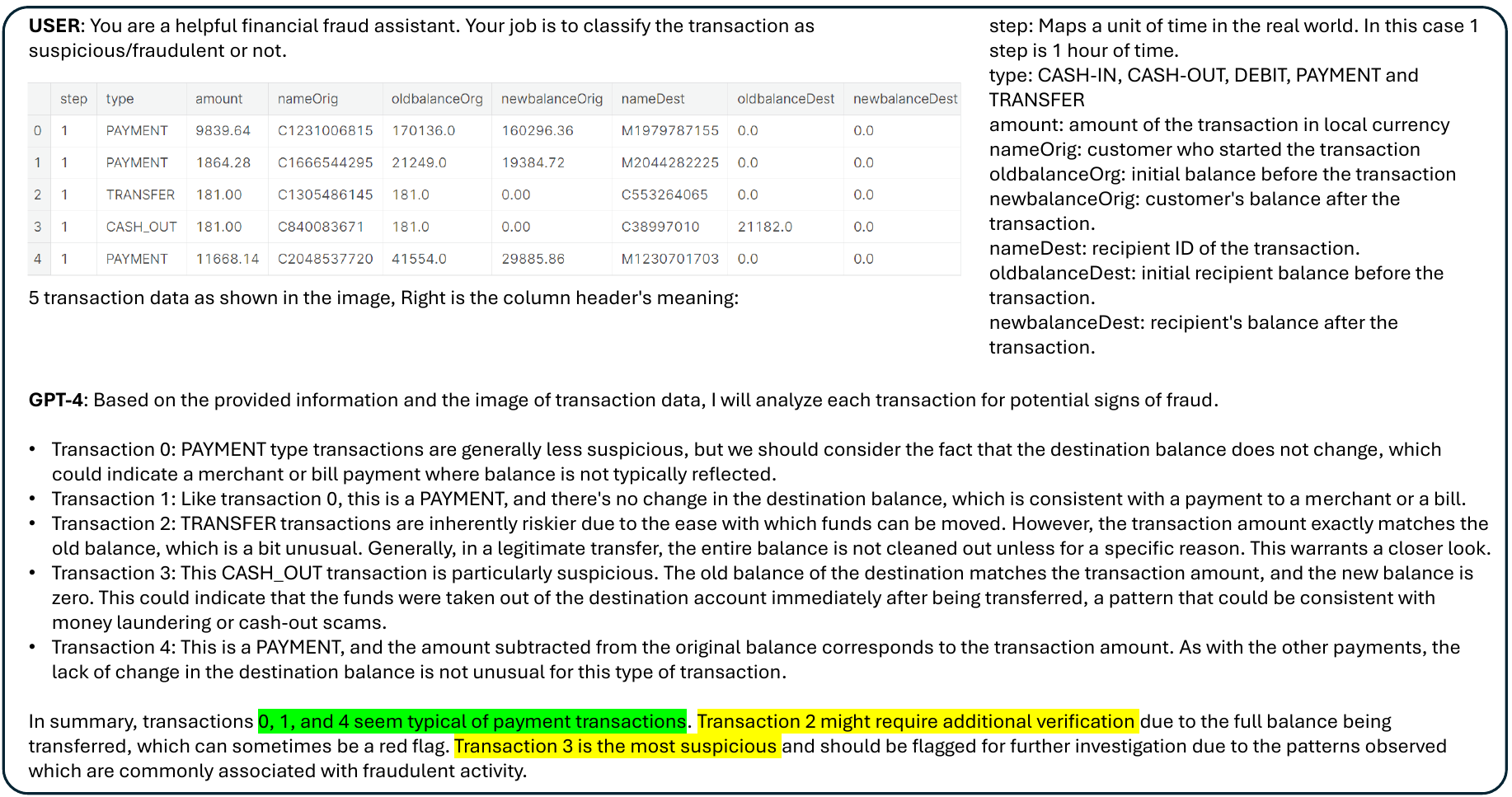}
\caption{Fraud detection on PaySim simulates mobile money transactions dataset  \cite{lopez2016paysim} using GPT-4: 5 out of 5 Correct. Green highlight indicates normal transactions; Yellow highlight indicates suspicious/fraudulent transactions.} 
\end{figure*}

\textbf{Evaluating Sentiment in Financial News:} The task of discerning sentiment in financial news is a paramount concern within the financial analytics community, as underscored in seminal works by Araci~ \cite{araci2019finbert} and Yang et al. ~ \cite{yang2020finbert}. This endeavor seeks to meticulously interpret the sentiment embedded in financial narratives. Adhering to the established FLUE framework~ \cite{shah2022flue}, this study employs two prominent datasets: the Financial Phrase Bank (FPB) dataset~ \cite{malo2014good} and FiQA-SA~ \cite{maia201818}. The FPB dataset is a collection of financial news excerpts, each meticulously annotated by field experts with a sentiment classification: positive, negative, or neutral. Conversely, FiQA-SA serves as an expansive dataset predominantly utilized for the sentiment quantification of English-language financial reporting and microblogging content, using a nuanced sentiment intensity scale ranging from -1 to 1, where a value of 1 epitomizes the most positive sentiment.

\textbf{Identifying Named Entities in Finance:} The goal of this task is to pinpoint key financial entities, including individuals, organizations, and locations. These entities are crucial for developing financial knowledge graphs. The NER dataset~ \cite{alvarado2015domain} serves as the basis for this task, featuring sentences from financial agreements filed with the U.S. Securities and Exchange Commission and includes entities categorized as LOCATION, ORGANISATION, and PERSON.

\textbf{Financial Question Answering:} This task involves automatically responding to financial queries based on provided data. For this, two datasets are employed: FinQA~ \cite{chen2021finqa} and ConvFinQA~ \cite{chen2022convfinqa}. FinQA offers pairs of questions and answers, annotated by specialists, along with associated earnings reports from S\&P 500 companies. ConvFinQA extends this by including multi-turn dialogues over these earnings reports.

\textbf{Predicting Stock Movements:} Recognized as a critical financial task, predicting stock movements can be invaluable in practical applications like investment strategy formulation. This task is approached as a binary classification challenge, following the methodology of prior research~ \cite{soun2022accurate}. It involves forecasting the direction of stock price movements based on historical prices and relevant tweets. Movements above 0.55\% are considered positive, while those below -0.5\% are deemed negative. For this analysis, one widely used datasets is utilized: BigData22~ \cite{soun2022accurate}.

\subsection{Prompt Design}
We examined various prompting strategies, including vanilla zero-shot prompting, Chain-of Thought (CoT) enhanced zero-shot prompting, and one-shot prompting to investigate their impact on GPT’s performance in the stated financial tasks.
The formulation of prompts is essential in interacting with LLMs. An elaborate and well-organized prompt, detailed and clear, leads to outputs that are more accurate and in line with the provided instructions. The following three parts are the GPT4-prompt in the financial field that we have obtained through experiments to generate output that best meets the instructions.
\begin{itemize}
\item \textbf{System Role Explanation}: This section will describe the specific role and tasks GPT-4 is expected to perform in a financial setting. For example, it might be tasked with analyzing market trends, offering investment advice, or interpreting financial reports.
\item \textbf{Response Format for Different Tasks}: This section has specific requirements for the format of the output. For example, you want the information to be presented in the form of lists, charts, or reports.
\item \textbf{Example and Output}: This part would provide a example as a guideline for a finance-related query. Also, it contains the desired response for the query.
\
\end{itemize}

The first three components are utilized for system message input in response to each query. We have attempted to enhance the precision of the prompts by incorporating additional components, yet this has not resulted in a substantial improvement in performance.

\section{Experimental Results}

In our tested financial tasks, LLMs demonstrated precise execution capabilities. Based on the responses we gathered, we believe that LLMs exhibit exceptional zero-shot learning and mathematical reasoning abilities, along with their strongest suit, language sentiment analysis. The effectiveness of LLMs in financial tasks is quantitatively assessed by comparing their recommendations against real-world financial data and historical market performance. This methodology was rigorously tested across various financial scenarios and datasets, yielding insightful and actionable results in areas such as financial engineering, risk assessment, and market trend analysis. For financial tasks lacking dedicated datasets, we have curated case studies to showcase the capabilities of GPT-4.

\begin{table}[ht]
\centering
\caption{The zero-shot and few-shot performance of different LLMs on the stated datasets.}
\begin{tabular}{@{}llc@{}}
\toprule
Dataset & Metrics & GPT 4 \\ 
\midrule
FPB & Acc & 0.78 \\
FiQA-SA & Acc & 0.79 \\
NER & EntityF1 & 0.81 \\
FinQA & EmAcc & 0.64 \\
ConvFinQA & EmAcc & 0.73\\
BigData22 & Acc & 0.53 \\

\bottomrule
\end{tabular}
\end{table}

\section{Limitation and Future work}

The limitations of LLMs are evident in areas such as optimization and quantitative trading. While they can assist in identifying market sentiments, LLMs cannot directly engage in computational tasks. Their role is more auxiliary, aiding in sentiment analysis which then feeds into existing models that handle quantitative variables \cite{wang2023alphagpt}. This indicates that LLMs, as of now, are not standalone solutions for computational finance tasks but rather powerful tools for augmenting existing models.

For future work, there is immense potential in integrating LLMs with advanced quantitative models. One promising direction could be the development of hybrid systems that combine the text processing prowess of LLMs with sophisticated quantitative trading algorithms \cite{yu2023finme,ciuriak2023trading}. Another area could be enhancing the interpretability and reliability of LLMs outputs in financial contexts, ensuring that the insights generated are not only accurate but also actionable. Moreover, exploring the application of LLMs in predictive analytics for market trends, based on historical data and current events, can open new avenues in financial forecasting. This integration of qualitative and quantitative analysis could revolutionize how financial markets are analyzed and traded \cite{ge2023openagi}.

\section{Conclusion}
In this article, we have delved into the multifaceted application of GPT-4 across a spectrum of 11 financial tasks, shedding light on the capabilities and constraints of LLMs in the financial domain. Central to our findings is the remarkable adeptness of LLMs in text processing, sentiment analysis, and their zero-shot learning abilities. The proficiency of LLMs in sifting through and interpreting extensive textual data is unmatched, thus playing a pivotal role in decoding market dynamics and investor sentiment.

However, it is crucial to acknowledge the limitations of LLMs in direct computational tasks, particularly in optimization and quantitative trading, where their role remains largely supplementary. Despite these constraints, the potential of LLMs in enhancing financial models and decision-making processes is undeniable. As we advance, the integration of LLMs with quantitative models and the refinement of their application in finance will be areas of significant interest. The continual evolution of LLMs promises to not only bolster existing financial methodologies but also to pave the way for innovative approaches in financial analysis and strategy.

\bibliographystyle{splncs04}
\bibliography{mybib}

@article{yang2023fingpt,
  title={FinGPT: Open-Source Financial Large Language Models},
  author={Yang, Hongyang and Liu, Xiao-Yang and Wang, Christina Dan},
  journal={arXiv preprint arXiv:2306.06031},
  year={2023}
}

@article{ge2023openagi,
  title={Openagi: When llm meets domain experts},
  author={Ge, Yingqiang and Hua, Wenyue and Ji, Jianchao and Tan, Juntao and Xu, Shuyuan and Zhang, Yongfeng},
  journal={arXiv preprint arXiv:2304.04370},
  year={2023}
}

@article{ciuriak2023trading,
  title={Trading AI: Machine Knowledge Capital and the Trading System},
  author={Ciuriak, Dan and Artyushina, Anna},
  journal={Available at SSRN},
  year={2023}
}

@article{yu2023finme,
  title={FinMe: A Performance-Enhanced Large Language Model Trading Agent with Layered Memory and Character Design},
  author={Yu, Yangyang and Li, Haohang and Chen, Zhi and Jiang, Yuechen and Li, Yang and Zhang, Denghui and Liu, Rong and Suchow, Jordan W and Khashanah, Khaldoun},
  journal={arXiv preprint arXiv:2311.13743},
  year={2023}
}

@article{kojima2022large,
  title={Large language models are zero-shot reasoners},
  author={Kojima, Takeshi and Gu, Shixiang Shane and Reid, Machel and Matsuo, Yutaka and Iwasawa, Yusuke},
  journal={Advances in neural information processing systems},
  volume={35},
  pages={22199--22213},
  year={2022}
}

@article{zheng2023llm,
  title={LLM-FuncMapper: Function Identification for Interpreting Complex Clauses in Building Codes via LLM},
  author={Zheng, Zhe and Chen, Ke-Yin and Cao, Xin-Yu and Lu, Xin-Zheng and Lin, Jia-Rui},
  journal={arXiv preprint arXiv:2308.08728},
  year={2023}
}

@article{liu2023gpt,
  title={A GPT-based method of Automated Compliance Checking through prompt engineering},
  author={Liu, Xiaoyu and Li, Haijiang and Zhu, Xiaofeng},
  year={2023}
}

@article{zheng2020effects,
  title={The effects of class imbalance and training data size on classifier learning: an empirical study},
  author={Zheng, Wanwan and Jin, Mingzhe},
  journal={SN Computer Science},
  volume={1},
  pages={1--13},
  year={2020},
  publisher={Springer}
}

@article{babaei2023gpt,
  title={GPT Classifications, With Application to Credit Lending},
  author={Babaei, Golnoosh and Giudici, Paolo},
  journal={Available at SSRN 4649285},
  year={2023}
}

@article{yoon2023design,
  title={Design and Implementation of an LLM system to Improve Response Time for SMEs Technology Credit Evaluation},
  author={Yoon, Sungwook},
  journal={International journal of advanced smart convergence},
  volume={12},
  number={3},
  pages={51--60},
  year={2023},
  publisher={The Institute of Internet, Broadcasting and Communication}
}

@article{aw2023counteracting,
  title={Counteracting dark sides of robo-advisors: justice, privacy and intrusion considerations},
  author={Aw, Eugene Cheng-Xi and Leong, Lai-Ying and Hew, Jun-Jie and Rana, Nripendra P and Tan, Teck Ming and Jee, Teck-Weng},
  journal={International Journal of Bank Marketing},
  year={2023},
  publisher={Emerald Publishing Limited}
}

@article{dastile2020statistical,
  title={Statistical and machine learning models in credit scoring: A systematic literature survey},
  author={Dastile, Xolani and Celik, Turgay and Potsane, Moshe},
  journal={Applied Soft Computing},
  volume={91},
  pages={106263},
  year={2020},
  publisher={Elsevier}
}

@article{scherer2023trust,
  title={Trust me, I am a Robo-advisor},
  author={Scherer, Bernd and Lehner, Sebastian},
  journal={Journal of Asset Management},
  volume={24},
  number={2},
  pages={85--96},
  year={2023},
  publisher={Springer}
}

@article{lourencco2020whose,
  title={Whose algorithm says so: The relationships between type of firm, perceptions of trust and expertise, and the acceptance of financial robo-advice},
  author={Louren{\c{c}}o, Carlos JS and Dellaert, Benedict GC and Donkers, Bas},
  journal={Journal of Interactive Marketing},
  volume={49},
  pages={107--124},
  year={2020},
  publisher={Elsevier}
}

@article{baker2017regulating,
  title={Regulating robo advice across the financial services industry},
  author={Baker, Tom and Dellaert, Benedict},
  journal={Iowa L. Rev.},
  volume={103},
  pages={713},
  year={2017},
  publisher={HeinOnline}
}

@article{caspigenerative,
  title={Generative AI and the Future of Financial Advice Regulation},
  author={Caspi, Itamar and Felber, Sarith S and Gillis, Talia B}
}

@article{aligeopolitical,
  title={Geopolitical Threat, Market Capitalization, and Portfolio Return},
  author={Ali, Syed Riaz Mahmood},
  journal={Market Capitalization, and Portfolio Return}
}

@article{paulson1991makes,
  title={What Makes a Portfolio a Portfolio?.},
  author={Paulson, F Leon and others},
  journal={Educational leadership},
  volume={48},
  number={5},
  pages={60--63},
  year={1991},
  publisher={ERIC}
}

@article{chang2023llm4ts,
  title={Llm4ts: Two-stage fine-tuning for time-series forecasting with pre-trained llms},
  author={Chang, Ching and Peng, Wen-Chih and Chen, Tien-Fu},
  journal={arXiv preprint arXiv:2308.08469},
  year={2023}
}

@article{jin2023time,
  title={Time-llm: Time series forecasting by reprogramming large language models},
  author={Jin, Ming and Wang, Shiyu and Ma, Lintao and Chu, Zhixuan and Zhang, James Y and Shi, Xiaoming and Chen, Pin-Yu and Liang, Yuxuan and Li, Yuan-Fang and Pan, Shirui and others},
  journal={arXiv preprint arXiv:2310.01728},
  year={2023}
}

@article{touvron2023llama,
  title={Llama: Open and efficient foundation language models},
  author={Touvron, Hugo and Lavril, Thibaut and Izacard, Gautier and Martinet, Xavier and Lachaux, Marie-Anne and Lacroix, Timoth{\'e}e and Rozi{\`e}re, Baptiste and Goyal, Naman and Hambro, Eric and Azhar, Faisal and others},
  journal={arXiv preprint arXiv:2302.13971},
  year={2023}
}

@article{achiam2023gpt,
  title={Gpt-4 technical report},
  author={Achiam, Josh and Adler, Steven and Agarwal, Sandhini and Ahmad, Lama and Akkaya, Ilge and Aleman, Florencia Leoni and Almeida, Diogo and Altenschmidt, Janko and Altman, Sam and Anadkat, Shyamal and others},
  journal={arXiv preprint arXiv:2303.08774},
  year={2023}
}

@article{brown2020language,
  title={Language models are few-shot learners},
  author={Brown, Tom and Mann, Benjamin and Ryder, Nick and Subbiah, Melanie and Kaplan, Jared D and Dhariwal, Prafulla and Neelakantan, Arvind and Shyam, Pranav and Sastry, Girish and Askell, Amanda and others},
  journal={Advances in neural information processing systems},
  volume={33},
  pages={1877--1901},
  year={2020}
}

@book{johansen1995likelihood,
  title={Likelihood-based inference in cointegrated vector autoregressive models},
  author={Johansen, S{\o}ren},
  year={1995},
  publisher={OUP Oxford}
}

@article{fan2005selective,
  title={A selective overview of nonparametric methods in financial econometrics},
  author={Fan, Jianqing},
  journal={Statistical Science},
  pages={317--337},
  year={2005},
  publisher={JSTOR}
}

@inproceedings{orderud2005comparison,
  title={Comparison of kalman filter estimation approaches for state space models with nonlinear measurements},
  author={Orderud, Fredrik},
  booktitle={Proc. of Scandinavian Conference on Simulation and Modeling},
  pages={1--8},
  year={2005}
}

@article{zivot2006vector,
  title={Vector autoregressive models for multivariate time series},
  author={Zivot, Eric and Wang, Jiahui},
  journal={Modeling financial time series with S-PLUS{\textregistered}},
  pages={385--429},
  year={2006},
  publisher={Springer}
}

@inproceedings{radha2006forecasting,
  title={Forecasting short term interest rates using ARMA, ARMA-GARCH and ARMA-EGARCH models},
  author={Radha, S and Thenmozhi, M},
  booktitle={Indian Institute of Capital Markets 9th Capital Markets Conference Paper},
  year={2006}
}

@inproceedings{zhao2021bert,
  title={A BERT based sentiment analysis and key entity detection approach for online financial texts},
  author={Zhao, Lingyun and Li, Lin and Zheng, Xinhao and Zhang, Jianwei},
  booktitle={2021 IEEE 24th International Conference on Computer Supported Cooperative Work in Design (CSCWD)},
  pages={1233--1238},
  year={2021},
  organization={IEEE}
}

@article{luo2018attention,
  title={An attention-based BiLSTM-CRF approach to document-level chemical named entity recognition},
  author={Luo, Ling and Yang, Zhihao and Yang, Pei and Zhang, Yin and Wang, Lei and Lin, Hongfei and Wang, Jian},
  journal={Bioinformatics},
  volume={34},
  number={8},
  pages={1381--1388},
  year={2018},
  publisher={Oxford University Press}
}

@article{ekbal2010named,
  title={Named entity recognition using support vector machine: A language independent approach},
  author={Ekbal, Asif and Bandyopadhyay, Sivaji},
  journal={International Journal of Electrical and Computer Engineering},
  volume={4},
  number={3},
  pages={589--604},
  year={2010}
}

@article{li2020survey,
  title={A survey on deep learning for named entity recognition},
  author={Li, Jing and Sun, Aixin and Han, Jianglei and Li, Chenliang},
  journal={IEEE Transactions on Knowledge and Data Engineering},
  volume={34},
  number={1},
  pages={50--70},
  year={2020},
  publisher={IEEE}
}

@article{nasar2021named,
  title={Named entity recognition and relation extraction: State-of-the-art},
  author={Nasar, Zara and Jaffry, Syed Waqar and Malik, Muhammad Kamran},
  journal={ACM Computing Surveys (CSUR)},
  volume={54},
  number={1},
  pages={1--39},
  year={2021},
  publisher={ACM New York, NY, USA}
}

@article{yang2020finbert,
  title={Finbert: A pretrained language model for financial communications},
  author={Yang, Yi and Uy, Mark Christopher Siy and Huang, Allen},
  journal={arXiv preprint arXiv:2006.08097},
  year={2020}
}

@misc{yu2023temporal,
      title={Temporal Data Meets LLM -- Explainable Financial Time Series Forecasting}, 
      author={Xinli Yu and Zheng Chen and Yuan Ling and Shujing Dong and Zongyi Liu and Yanbin Lu},
      year={2023},
      eprint={2306.11025},
      archivePrefix={arXiv},
      primaryClass={cs.LG}
}

@misc{wu2023bloomberggpt,
      title={BloombergGPT: A Large Language Model for Finance}, 
      author={Shijie Wu and Ozan Irsoy and Steven Lu and Vadim Dabravolski and Mark Dredze and Sebastian Gehrmann and Prabhanjan Kambadur and David Rosenberg and Gideon Mann},
      year={2023},
      eprint={2303.17564},
      archivePrefix={arXiv},
      primaryClass={cs.LG}
}

@misc{liu2023fingpt,
      title={FinGPT: Democratizing Internet-scale Data for Financial Large Language Models}, 
      author={Xiao-Yang Liu and Guoxuan Wang and Hongyang Yang and Daochen Zha},
      year={2023},
      eprint={2307.10485},
      archivePrefix={arXiv},
      primaryClass={cs.CL}
}

@misc{xie2023pixiu,
      title={PIXIU: A Large Language Model, Instruction Data and Evaluation Benchmark for Finance}, 
      author={Qianqian Xie and Weiguang Han and Xiao Zhang and Yanzhao Lai and Min Peng and Alejandro Lopez-Lira and Jimin Huang},
      year={2023},
      eprint={2306.05443},
      archivePrefix={arXiv},
      primaryClass={cs.CL}
}

@inproceedings{soun2022accurate,
  title={Accurate Stock Movement Prediction with Self-supervised Learning from Sparse Noisy Tweets},
  author={Soun, Yejun and Yoo, Jaemin and Cho, Minyong and Jeon, Jihyeong and Kang, U},
  booktitle={2022 IEEE International Conference on Big Data (Big Data)},
  pages={1691--1700},
  year={2022},
  organization={IEEE}
}

@article{ehrmann2023named,
  title={Named entity recognition and classification in historical documents: A survey},
  author={Ehrmann, Maud and Hamdi, Ahmed and Pontes, Elvys Linhares and Romanello, Matteo and Doucet, Antoine},
  journal={ACM Computing Surveys},
  volume={56},
  number={2},
  pages={1--47},
  year={2023},
  publisher={ACM New York, NY}
}

@misc{ehrmann2021named,
      title={Named Entity Recognition and Classification on Historical Documents: A Survey}, 
      author={Maud Ehrmann and Ahmed Hamdi and Elvys Linhares Pontes and Matteo Romanello and Antoine Doucet},
      year={2021},
      eprint={2109.11406},
      archivePrefix={arXiv},
      primaryClass={cs.CL}
}

@article{bollen2011twitter,
  title={Twitter mood predicts the stock market},
  author={Bollen, Johan and Mao, Huina and Zeng, Xiaojun},
  journal={Journal of computational science},
  volume={2},
  number={1},
  pages={1--8},
  year={2011},
  publisher={Elsevier}
}

@article{li2014effect,
  title={The effect of news and public mood on stock movements},
  author={Li, Qing and Wang, TieJun and Li, Ping and Liu, Ling and Gong, Qixu and Chen, Yuanzhu},
  journal={Information Sciences},
  volume={278},
  pages={826--840},
  year={2014},
  publisher={Elsevier}
}

@article{yu2019information,
  title={Information availability and return volatility in the bitcoin market: analyzing differences of user opinion and interest},
  author={Yu, Ju Hyun and Kang, Juyoung and Park, Sangun},
  journal={Information Processing \& Management},
  volume={56},
  number={3},
  pages={721--732},
  year={2019},
  publisher={Elsevier}
}

@inproceedings{shah2018predicting,
  title={Predicting the effects of news sentiments on the stock market},
  author={Shah, Dev and Isah, Haruna and Zulkernine, Farhana},
  booktitle={2018 IEEE International Conference on Big Data (Big Data)},
  pages={4705--4708},
  year={2018},
  organization={IEEE}
}

@inproceedings{usha2013analysis,
  title={Analysis of sentiments using unsupervised learning techniques},
  author={Usha, MS and Devi, M Indra},
  booktitle={2013 International Conference on Information Communication and Embedded Systems (ICICES)},
  pages={241--245},
  year={2013},
  organization={IEEE}
}

@article{kirange2016sentiment,
  title={Sentiment Analysis of news headlines for stock price prediction},
  author={Kirange, DK and Deshmukh, Ratnadeep R and others},
  journal={Composoft, An International Journal of Advanced Computer Technology},
  volume={5},
  number={3},
  pages={2080--2084},
  year={2016}
}

@inproceedings{kalra2019efficacy,
  title={Efficacy of news sentiment for stock market prediction},
  author={Kalra, Sneh and Prasad, Jay Shankar},
  booktitle={2019 International Conference on Machine Learning, Big Data, Cloud and Parallel Computing (COMITCon)},
  pages={491--496},
  year={2019},
  organization={IEEE}
}

@article{li2020incorporating,
  title={Incorporating stock prices and news sentiments for stock market prediction: A case of Hong Kong},
  author={Li, Xiaodong and Wu, Pangjing and Wang, Wenpeng},
  journal={Information Processing \& Management},
  volume={57},
  number={5},
  pages={102212},
  year={2020},
  publisher={Elsevier}
}

@article{sert2020analysis,
  title={Analysis and prediction in sparse and high dimensional text data: The case of Dow Jones stock market},
  author={Sert, Onur Can and {\c{S}}ahin, Salih Doruk and {\"O}zyer, Tansel and Alhajj, Reda},
  journal={Physica A: Statistical Mechanics and its Applications},
  volume={545},
  pages={123752},
  year={2020},
  publisher={Elsevier}
}

@article{nguyen2015sentiment,
  title={Sentiment analysis on social media for stock movement prediction},
  author={Nguyen, Thien Hai and Shirai, Kiyoaki and Velcin, Julien},
  journal={Expert Systems with Applications},
  volume={42},
  number={24},
  pages={9603--9611},
  year={2015},
  publisher={Elsevier}
}

@article{khedr2017predicting,
  title={Predicting stock market behavior using data mining technique and news sentiment analysis},
  author={Khedr, Ayman E and Yaseen, Nagwa and others},
  journal={International Journal of Intelligent Systems and Applications},
  volume={9},
  number={7},
  pages={22},
  year={2017},
  publisher={Modern Education and Computer Science Press}
}

@inproceedings{kalyanaraman2014sentiment,
  title={Sentiment analysis on news articles for stocks},
  author={Kalyanaraman, Vaanchitha and Kazi, Sarah and Tondulkar, Rohan and Oswal, Sangeeta},
  booktitle={2014 8th Asia Modelling Symposium},
  pages={10--15},
  year={2014},
  organization={IEEE}
}

@article{mittal2012stock,
  title={Stock prediction using twitter sentiment analysis},
  author={Mittal, Anshul and Goel, Arpit},
  journal={Standford University, CS229 (2011 http://cs229. stanford. edu/proj2011/GoelMittal-StockMarketPredictionUsingTwitterSentimentAnalysis. pdf)},
  volume={15},
  pages={2352},
  year={2012}
}

@article{audrino2020impact,
  title={The impact of sentiment and attention measures on stock market volatility},
  author={Audrino, Francesco and Sigrist, Fabio and Ballinari, Daniele},
  journal={International Journal of Forecasting},
  volume={36},
  number={2},
  pages={334--357},
  year={2020},
  publisher={Elsevier}
}

@inproceedings{mohan2019stock,
  title={Stock price prediction using news sentiment analysis},
  author={Mohan, Saloni and Mullapudi, Sahitya and Sammeta, Sudheer and Vijayvergia, Parag and Anastasiu, David C},
  booktitle={2019 IEEE fifth international conference on big data computing service and applications (BigDataService)},
  pages={205--208},
  year={2019},
  organization={IEEE}
}

@article{clement2022improving,
  title={Improving ESG scores with sustainability concepts},
  author={Cl{\'e}ment, Alexandre and Robinot, {\'E}lisabeth and Trespeuch, L{\'e}o},
  journal={Sustainability},
  volume={14},
  number={20},
  pages={13154},
  year={2022},
  publisher={MDPI}
}

@article{escrig2019rating,
  title={Rating the raters: Evaluating how ESG rating agencies integrate sustainability principles},
  author={Escrig-Olmedo, Elena and Fern{\'a}ndez-Izquierdo, Mar{\'\i}a {\'A}ngeles and Ferrero-Ferrero, Idoya and Rivera-Lirio, Juana Mar{\'\i}a and Mu{\~n}oz-Torres, Mar{\'\i}a Jes{\'u}s},
  journal={Sustainability},
  volume={11},
  number={3},
  pages={915},
  year={2019},
  publisher={Mdpi}
}

@article{friede2015esg,
  title={ESG and financial performance: aggregated evidence from more than 2000 empirical studies},
  author={Friede, Gunnar and Busch, Timo and Bassen, Alexander},
  journal={Journal of sustainable finance \& investment},
  volume={5},
  number={4},
  pages={210--233},
  year={2015},
  publisher={Taylor \& Francis}
}

@misc{zumente2021esg,
  title={ESG rating—necessity for the investor or the company? Sustainability, 13 (16), 8940},
  author={Zumente, I and L{\=a}ce, N},
  year={2021}
}

@article{liu2023transformation,
  title={Transformation vs Tradition: Artificial General Intelligence (AGI) for Arts and Humanities},
  author={Liu, Zhengliang and Li, Yiwei and Cao, Qian and Chen, Junwen and Yang, Tianze and Wu, Zihao and Hale, John and Gibbs, John and Rasheed, Khaled and Liu, Ninghao and others},
  journal={arXiv preprint arXiv:2310.19626},
  year={2023}
}

@inproceedings{rezayi2022agribert,
  title={Agribert: knowledge-infused agricultural language models for matching food and nutrition},
  author={Rezayi, Saed and Liu, Zhengliang and Wu, Zihao and Dhakal, Chandra and Ge, Bao and Zhen, Chen and Liu, Tianming and Li, Sheng},
  booktitle={Proceedings of the Thirty-First International Joint Conference on Artificial Intelligence},
  volume={7},
  pages={5150--5156},
  year={2022}
}

@article{rezayi2023exploring,
  title={Exploring New Frontiers in Agricultural NLP: Investigating the Potential of Large Language Models for Food Applications},
  author={Rezayi, Saed and Liu, Zhengliang and Wu, Zihao and Dhakal, Chandra and Ge, Bao and Dai, Haixing and Mai, Gengchen and Liu, Ninghao and Zhen, Chen and Liu, Tianming and others},
  journal={arXiv preprint arXiv:2306.11892},
  year={2023}
}

@article{freitag2017beam,
  title={Beam search strategies for neural machine translation},
  author={Freitag, Markus and Al-Onaizan, Yaser},
  journal={arXiv preprint arXiv:1702.01806},
  year={2017}
}

@article{tang2023science,
  title={The science of detecting llm-generated texts},
  author={Tang, Ruixiang and Chuang, Yu-Neng and Hu, Xia},
  journal={arXiv preprint arXiv:2303.07205},
  year={2023}
}

@article{zhao2023brain,
  title={When brain-inspired ai meets agi},
  author={Zhao, Lin and Zhang, Lu and Wu, Zihao and Chen, Yuzhong and Dai, Haixing and Yu, Xiaowei and Liu, Zhengliang and Zhang, Tuo and Hu, Xintao and Jiang, Xi and others},
  journal={Meta-Radiology},
  pages={100005},
  year={2023},
  publisher={Elsevier}
}

@article{vaswani2017attention,
  title={Attention is all you need},
  author={Vaswani, Ashish and Shazeer, Noam and Parmar, Niki and Uszkoreit, Jakob and Jones, Llion and Gomez, Aidan N and Kaiser, {\L}ukasz and Polosukhin, Illia},
  journal={Advances in neural information processing systems},
  volume={30},
  year={2017}
}

@article{zhao2023survey,
  title={A survey of large language models},
  author={Zhao, Wayne Xin and Zhou, Kun and Li, Junyi and Tang, Tianyi and Wang, Xiaolei and Hou, Yupeng and Min, Yingqian and Zhang, Beichen and Zhang, Junjie and Dong, Zican and others},
  journal={arXiv preprint arXiv:2303.18223},
  year={2023}
}

@article{chang2023survey,
  title={A survey on evaluation of large language models},
  author={Chang, Yupeng and Wang, Xu and Wang, Jindong and Wu, Yuan and Zhu, Kaijie and Chen, Hao and Yang, Linyi and Yi, Xiaoyuan and Wang, Cunxiang and Wang, Yidong and others},
  journal={arXiv preprint arXiv:2307.03109},
  year={2023}
}

@article{liu2023summary,
  title={Summary of ChatGPT-Related Research and Perspective Towards the Future of Large Language Models},
  author={Liu, Yiheng and Han, Tianle and Ma, Siyuan and Zhang, Jiayue and Yang, Yuanyuan and Tian, Jiaming and He, Hao and Li, Antong and He, Mengshen and Liu, Zhengliang and others},
  journal={Meta-Radiology},
  pages={100017},
  year={2023},
  publisher={Elsevier}
}

@article{lewis2019bart,
  title={Bart: Denoising sequence-to-sequence pre-training for natural language generation, translation, and comprehension},
  author={Lewis, Mike and Liu, Yinhan and Goyal, Naman and Ghazvininejad, Marjan and Mohamed, Abdelrahman and Levy, Omer and Stoyanov, Ves and Zettlemoyer, Luke},
  journal={arXiv preprint arXiv:1910.13461},
  year={2019}
}

@article{raffel2020exploring,
  title={Exploring the limits of transfer learning with a unified text-to-text transformer},
  author={Raffel, Colin and Shazeer, Noam and Roberts, Adam and Lee, Katherine and Narang, Sharan and Matena, Michael and Zhou, Yanqi and Li, Wei and Liu, Peter J},
  journal={The Journal of Machine Learning Research},
  volume={21},
  number={1},
  pages={5485--5551},
  year={2020},
  publisher={JMLRORG}
}

@article{radford2018improving,
  title={Improving language understanding by generative pre-training},
  author={Radford, Alec and Narasimhan, Karthik and Salimans, Tim and Sutskever, Ilya and others},
  year={2018},
  publisher={OpenAI}
}

@article{maple2023ai,
  title={The AI revolution: opportunities and challenges for the finance sector},
  author={Maple, Carsten and Szpruch, Lukasz and Epiphaniou, Gregory and Staykova, Kalina and Singh, Simran and Penwarden, William and Wen, Yisi and Wang, Zijian and Hariharan, Jagdish and Avramovic, Pavle},
  journal={arXiv preprint arXiv:2308.16538},
  year={2023}
}

@article{nourallah2023one,
  title={One size does not fit all: Young retail investors’ initial trust in financial robo-advisors},
  author={Nourallah, Mustafa},
  journal={Journal of Business Research},
  volume={156},
  pages={113470},
  year={2023},
  publisher={Elsevier}
}

@misc{jeong2024finetuning,
      title={Fine-tuning and Utilization Methods of Domain-specific LLMs}, 
      author={Cheonsu Jeong},
      year={2024},
      eprint={2401.02981},
      archivePrefix={arXiv},
      primaryClass={cs.CL}
}

@misc{wang2023alphagpt,
      title={Alpha-GPT: Human-AI Interactive Alpha Mining for Quantitative Investment}, 
      author={Saizhuo Wang and Hang Yuan and Leon Zhou and Lionel M. Ni and Heung-Yeung Shum and Jian Guo},
      year={2023},
      eprint={2308.00016},
      archivePrefix={arXiv},
      primaryClass={q-fin.CP}
}

@inproceedings{rezayi2022clinicalradiobert,
  title={Clinicalradiobert: Knowledge-infused few shot learning for clinical notes named entity recognition},
  author={Rezayi, Saed and Dai, Haixing and Liu, Zhengliang and Wu, Zihao and Hebbar, Akarsh and Burns, Andrew H and Zhao, Lin and Zhu, Dajiang and Li, Quanzheng and Liu, Wei and others},
  booktitle={International Workshop on Machine Learning in Medical Imaging},
  pages={269--278},
  year={2022},
  organization={Springer}
}

@misc{liao2023maskguided,
      title={Mask-guided BERT for Few Shot Text Classification}, 
      author={Wenxiong Liao and Zhengliang Liu and Haixing Dai and Zihao Wu and Yiyang Zhang and Xiaoke Huang and Yuzhong Chen and Xi Jiang and Wei Liu and Dajiang Zhu and Tianming Liu and Sheng Li and Xiang Li and Hongmin Cai},
      year={2023},
      eprint={2302.10447},
      archivePrefix={arXiv},
      primaryClass={cs.CL}
}

@misc{dai2023auggpt,
      title={AugGPT: Leveraging ChatGPT for Text Data Augmentation}, 
      author={Haixing Dai and Zhengliang Liu and Wenxiong Liao and Xiaoke Huang and Yihan Cao and Zihao Wu and Lin Zhao and Shaochen Xu and Wei Liu and Ninghao Liu and Sheng Li and Dajiang Zhu and Hongmin Cai and Lichao Sun and Quanzheng Li and Dinggang Shen and Tianming Liu and Xiang Li},
      year={2023},
      eprint={2302.13007},
      archivePrefix={arXiv},
      primaryClass={cs.CL}
}

@misc{liu2023deidgpt,
      title={DeID-GPT: Zero-shot Medical Text De-Identification by GPT-4}, 
      author={Zhengliang Liu and Xiaowei Yu and Lu Zhang and Zihao Wu and Chao Cao and Haixing Dai and Lin Zhao and Wei Liu and Dinggang Shen and Quanzheng Li and Tianming Liu and Dajiang Zhu and Xiang Li},
      year={2023},
      eprint={2303.11032},
      archivePrefix={arXiv},
      primaryClass={cs.CL}
}

@misc{ma2023impressiongpt,
      title={ImpressionGPT: An Iterative Optimizing Framework for Radiology Report Summarization with ChatGPT}, 
      author={Chong Ma and Zihao Wu and Jiaqi Wang and Shaochen Xu and Yaonai Wei and Zhengliang Liu and Xi Jiang and Lei Guo and Xiaoyan Cai and Shu Zhang and Tuo Zhang and Dajiang Zhu and Dinggang Shen and Tianming Liu and Xiang Li},
      year={2023},
      eprint={2304.08448},
      archivePrefix={arXiv},
      primaryClass={cs.CL}
}

@misc{liao2023differentiate,
      title={Differentiate ChatGPT-generated and Human-written Medical Texts}, 
      author={Wenxiong Liao and Zhengliang Liu and Haixing Dai and Shaochen Xu and Zihao Wu and Yiyang Zhang and Xiaoke Huang and Dajiang Zhu and Hongmin Cai and Tianming Liu and Xiang Li},
      year={2023},
      eprint={2304.11567},
      archivePrefix={arXiv},
      primaryClass={cs.CL}
}

@misc{dai2023adautogpt,
      title={AD-AutoGPT: An Autonomous GPT for Alzheimer's Disease Infodemiology}, 
      author={Haixing Dai and Yiwei Li and Zhengliang Liu and Lin Zhao and Zihao Wu and Suhang Song and Ye Shen and Dajiang Zhu and Xiang Li and Sheng Li and Xiaobai Yao and Lu Shi and Quanzheng Li and Zhuo Chen and Donglan Zhang and Gengchen Mai and Tianming Liu},
      year={2023},
      eprint={2306.10095},
      archivePrefix={arXiv},
      primaryClass={cs.CL}
}

@misc{guan2023cohortgpt,
      title={CohortGPT: An Enhanced GPT for Participant Recruitment in Clinical Study}, 
      author={Zihan Guan and Zihao Wu and Zhengliang Liu and Dufan Wu and Hui Ren and Quanzheng Li and Xiang Li and Ninghao Liu},
      year={2023},
      eprint={2307.11346},
      archivePrefix={arXiv},
      primaryClass={cs.CL}
}

@misc{liu2023pharmacygpt,
      title={PharmacyGPT: The AI Pharmacist}, 
      author={Zhengliang Liu and Zihao Wu and Mengxuan Hu and Bokai Zhao and Lin Zhao and Tianyi Zhang and Haixing Dai and Xianyan Chen and Ye Shen and Sheng Li and Brian Murray and Tianming Liu and Andrea Sikora},
      year={2023},
      eprint={2307.10432},
      archivePrefix={arXiv},
      primaryClass={cs.CL}
}

@misc{wu2023exploring,
      title={Exploring the Trade-Offs: Unified Large Language Models vs Local Fine-Tuned Models for Highly-Specific Radiology NLI Task}, 
      author={Zihao Wu and Lu Zhang and Chao Cao and Xiaowei Yu and Haixing Dai and Chong Ma and Zhengliang Liu and Lin Zhao and Gang Li and Wei Liu and Quanzheng Li and Dinggang Shen and Xiang Li and Dajiang Zhu and Tianming Liu},
      year={2023},
      eprint={2304.09138},
      archivePrefix={arXiv},
      primaryClass={cs.CL}
}

@misc{liu2023radiologygpt,
      title={Radiology-GPT: A Large Language Model for Radiology}, 
      author={Zhengliang Liu and Aoxiao Zhong and Yiwei Li and Longtao Yang and Chao Ju and Zihao Wu and Chong Ma and Peng Shu and Cheng Chen and Sekeun Kim and Haixing Dai and Lin Zhao and Dajiang Zhu and Jun Liu and Wei Liu and Dinggang Shen and Xiang Li and Quanzheng Li and Tianming Liu},
      year={2023},
      eprint={2306.08666},
      archivePrefix={arXiv},
      primaryClass={cs.CL}
}

@misc{wang2023review,
      title={Review of Large Vision Models and Visual Prompt Engineering}, 
      author={Jiaqi Wang and Zhengliang Liu and Lin Zhao and Zihao Wu and Chong Ma and Sigang Yu and Haixing Dai and Qiushi Yang and Yiheng Liu and Songyao Zhang and Enze Shi and Yi Pan and Tuo Zhang and Dajiang Zhu and Xiang Li and Xi Jiang and Bao Ge and Yixuan Yuan and Dinggang Shen and Tianming Liu and Shu Zhang},
      year={2023},
      eprint={2307.00855},
      archivePrefix={arXiv},
      primaryClass={cs.CV}
}

@misc{li2023artificial,
      title={Artificial General Intelligence for Medical Imaging}, 
      author={Xiang Li and Lu Zhang and Zihao Wu and Zhengliang Liu and Lin Zhao and Yixuan Yuan and Jun Liu and Gang Li and Dajiang Zhu and Pingkun Yan and Quanzheng Li and Wei Liu and Tianming Liu and Dinggang Shen},
      year={2023},
      eprint={2306.05480},
      archivePrefix={arXiv},
      primaryClass={cs.AI}
}

@misc{cai2023coarsetofine,
      title={Coarse-to-fine Knowledge Graph Domain Adaptation based on Distantly-supervised Iterative Training}, 
      author={Hongmin Cai and Wenxiong Liao and Zhengliang Liu and Yiyang Zhang and Xiaoke Huang and Siqi Ding and Hui Ren and Zihao Wu and Haixing Dai and Sheng Li and Lingfei Wu and Ninghao Liu and Quanzheng Li and Tianming Liu and Xiang Li},
      year={2023},
      eprint={2211.02849},
      archivePrefix={arXiv},
      primaryClass={cs.AI}
}

@article{holmes2023evaluating,
  title={Evaluating Large Language Models on a Highly-specialized Topic},
  author={Holmes, J and Liu, Z and Zhang, L and Ding, Y and Sio, TT and McGee, LA and Ashman, JB and Li, X and Liu, T and Shen, J and others},
  journal={Radiation Oncology Physics},
  year={2023}
}

@InProceedings{10.1007/978-3-031-45673-2_46,
author="Liu, Zhengliang
and Zhong, Aoxiao
and Li, Yiwei
and Yang, Longtao
and Ju, Chao
and Wu, Zihao
and Ma, Chong
and Shu, Peng
and Chen, Cheng
and Kim, Sekeun
and Dai, Haixing
and Zhao, Lin
and Zhu, Dajiang
and Liu, Jun
and Liu, Wei
and Shen, Dinggang
and Li, Quanzheng
and Liu, Tianming
and Li, Xiang",
editor="Cao, Xiaohuan
and Xu, Xuanang
and Rekik, Islem
and Cui, Zhiming
and Ouyang, Xi",
title="Tailoring Large Language Models to Radiology: A Preliminary Approach to LLM Adaptation for a Highly Specialized Domain",
booktitle="Machine Learning in Medical Imaging",
year="2024",
publisher="Springer Nature Switzerland",
address="Cham",
pages="464--473",
isbn="978-3-031-45673-2"
}

@misc{liu2023radiologyllama2,
      title={Radiology-Llama2: Best-in-Class Large Language Model for Radiology}, 
      author={Zhengliang Liu and Yiwei Li and Peng Shu and Aoxiao Zhong and Longtao Yang and Chao Ju and Zihao Wu and Chong Ma and Jie Luo and Cheng Chen and Sekeun Kim and Jiang Hu and Haixing Dai and Lin Zhao and Dajiang Zhu and Jun Liu and Wei Liu and Dinggang Shen and Tianming Liu and Quanzheng Li and Xiang Li},
      year={2023},
      eprint={2309.06419},
      archivePrefix={arXiv},
      primaryClass={cs.CL}
}

@misc{tang2023policygpt,
      title={PolicyGPT: Automated Analysis of Privacy Policies with Large Language Models}, 
      author={Chenhao Tang and Zhengliang Liu and Chong Ma and Zihao Wu and Yiwei Li and Wei Liu and Dajiang Zhu and Quanzheng Li and Xiang Li and Tianming Liu and Lei Fan},
      year={2023},
      eprint={2309.10238},
      archivePrefix={arXiv},
      primaryClass={cs.CL}
}

@misc{liu2023radoncgpt,
      title={RadOnc-GPT: A Large Language Model for Radiation Oncology}, 
      author={Zhengliang Liu and Peilong Wang and Yiwei Li and Jason Holmes and Peng Shu and Lian Zhang and Chenbin Liu and Ninghao Liu and Dajiang Zhu and Xiang Li and Quanzheng Li and Samir H. Patel and Terence T. Sio and Tianming Liu and Wei Liu},
      year={2023},
      eprint={2309.10160},
      archivePrefix={arXiv},
      primaryClass={physics.med-ph}
}

@misc{liuradiology,
  title={Radiology-GPT: a large language model for radiology. arXiv [Preprint]. 2023 [cited August 21, 2023]},
  author={Liu, Z and Zhong, A and Li, Y and Yang, L and Ju, C and Wu, Z and others}
}

@article{LIU2023100045,
title = {Artificial General Intelligence for Radiation Oncology},
journal = {Meta-Radiology},
pages = {100045},
year = {2023},
issn = {2950-1628},
doi = {https://doi.org/10.1016/j.metrad.2023.100045},
url = {https://www.sciencedirect.com/science/article/pii/S2950162823000450},
author = {Chenbin Liu and Zhengliang Liu and Jason Holmes and Lu Zhang and Lian Zhang and Yuzhen Ding and Peng Shu and Zihao Wu and Haixing Dai and Yiwei Li and Dinggang Shen and Ninghao Liu and Quanzheng Li and Xiang Li and Dajiang Zhu and Tianming Liu and Wei Liu}
}

@misc{dou2023artificial,
      title={Towards Artificial General Intelligence (AGI) in the Internet of Things (IoT): Opportunities and Challenges}, 
      author={Fei Dou and Jin Ye and Geng Yuan and Qin Lu and Wei Niu and Haijian Sun and Le Guan and Guoyu Lu and Gengchen Mai and Ninghao Liu and Jin Lu and Zhengliang Liu and Zihao Wu and Chenjiao Tan and Shaochen Xu and Xianqiao Wang and Guoming Li and Lilong Chai and Sheng Li and Jin Sun and Hongyue Sun and Yunli Shao and Changying Li and Tianming Liu and Wenzhan Song},
      year={2023},
      eprint={2309.07438},
      archivePrefix={arXiv},
      primaryClass={cs.AI}
}

@misc{gong2023evaluating,
      title={Evaluating the Potential of Leading Large Language Models in Reasoning Biology Questions}, 
      author={Xinyu Gong and Jason Holmes and Yiwei Li and Zhengliang Liu and Qi Gan and Zihao Wu and Jianli Zhang and Yusong Zou and Yuxi Teng and Tian Jiang and Hongtu Zhu and Wei Liu and Tianming Liu and Yajun Yan},
      year={2023},
      eprint={2311.07582},
      archivePrefix={arXiv},
      primaryClass={cs.CL}
}

@misc{holmes2023benchmarking,
      title={Benchmarking a foundation LLM on its ability to re-label structure names in accordance with the AAPM TG-263 report}, 
      author={Jason Holmes and Lian Zhang and Yuzhen Ding and Hongying Feng and Zhengliang Liu and Tianming Liu and William W. Wong and Sujay A. Vora and Jonathan B. Ashman and Wei Liu},
      year={2023},
      eprint={2310.03874},
      archivePrefix={arXiv},
      primaryClass={physics.med-ph}
}

@misc{shi2023mededit,
      title={MedEdit: Model Editing for Medical Question Answering with External Knowledge Bases}, 
      author={Yucheng Shi and Shaochen Xu and Zhengliang Liu and Tianming Liu and Xiang Li and Ninghao Liu},
      year={2023},
      eprint={2309.16035},
      archivePrefix={arXiv},
      primaryClass={cs.CL}
}

@article{liu2023artificial,
  title={Artificial General Intelligence for Radiation Oncology},
  author={Liu, Chenbin and Liu, Zhengliang and Holmes, Jason and Zhang, Lu and Zhang, Lian and Ding, Yuzhen and Shu, Peng and Wu, Zihao and Dai, Haixing and Li, Yiwei and others},
  journal={Meta-Radiology},
  pages={100045},
  year={2023},
  publisher={Elsevier}
}

@article{cai2022coarse,
  title={Coarse-to-fine knowledge graph domain adaptation based on distantly-supervised iterative training},
  author={Cai, Hongmin and Liao, Wenxiong and Liu, Zhengliang and Zhang, Yiyang and Huang, Xiaoke and Ding, Siqi and Ren, Hui and Wu, Zihao and Dai, Haixing and Li, Sheng and others},
  journal={arXiv preprint arXiv:2211.02849},
  year={2022}
}

@inproceedings{zhou2023fine,
  title={Fine-grained artificial neurons in audio-transformers for disentangling neural auditory encoding},
  author={Zhou, Mengyue and Liu, Xu and Liu, David and Wu, Zihao and Liu, Zhengliang and Zhao, Lin and Zhu, Dajiang and Guo, Lei and Han, Junwei and Liu, Tianming and others},
  booktitle={Findings of the Association for Computational Linguistics: ACL 2023},
  pages={7943--7956},
  year={2023}
}

@article{zhong2023chatabl,
  title={Chatabl: Abductive learning via natural language interaction with chatgpt},
  author={Zhong, Tianyang and Wei, Yaonai and Yang, Li and Wu, Zihao and Liu, Zhengliang and Wei, Xiaozheng and Li, Wenjun and Yao, Junjie and Ma, Chong and Li, Xiang and others},
  journal={arXiv preprint arXiv:2304.11107},
  year={2023}
}

@article{liu2022survey,
  title={Survey on natural language processing in medical image analysis.},
  author={Liu, Zhengliang and He, Mengshen and Jiang, Zuowei and Wu, Zihao and Dai, Haixing and Zhang, Lian and Luo, Siyi and Han, Tianle and Li, Xiang and Jiang, Xi and others},
  journal={Zhong nan da xue xue bao. Yi xue ban= Journal of Central South University. Medical Sciences},
  volume={47},
  number={8},
  pages={981--993},
  year={2022}
}

@article{liao2023mask,
  title={Mask-guided bert for few shot text classification},
  author={Liao, Wenxiong and Liu, Zhengliang and Dai, Haixing and Wu, Zihao and Zhang, Yiyang and Huang, Xiaoke and Chen, Yuzhong and Jiang, Xi and Zhu, Dajiang and Liu, Tianming and others},
  journal={arXiv preprint arXiv:2302.10447},
  year={2023}
}

@article{zhao2023ophtha,
  title={Ophtha-LLaMA2: A Large Language Model for Ophthalmology},
  author={Zhao, Huan and Ling, Qian and Pan, Yi and Zhong, Tianyang and Hu, Jin-Yu and Yao, Junjie and Xiao, Fengqian and Xiao, Zhenxiang and Zhang, Yutong and Xu, San-Hua and others},
  journal={arXiv preprint arXiv:2312.04906},
  year={2023}
}

@article{lee2023multimodality,
  title={Multimodality of AI for Education: Towards Artificial General Intelligence},
  author={Lee, Gyeong-Geon and Shi, Lehong and Latif, Ehsan and Gao, Yizhu and Bewersdorf, Arne and Nyaaba, Matthew and Guo, Shuchen and Wu, Zihao and Liu, Zhengliang and Wang, Hui and others},
  journal={arXiv preprint arXiv:2312.06037},
  year={2023}
}

@article{wei2022chain,
  title={Chain-of-thought prompting elicits reasoning in large language models},
  author={Wei, Jason and Wang, Xuezhi and Schuurmans, Dale and Bosma, Maarten and Xia, Fei and Chi, Ed and Le, Quoc V and Zhou, Denny and others},
  journal={Advances in Neural Information Processing Systems},
  volume={35},
  pages={24824--24837},
  year={2022}
}

@article{tan2023promises,
  title={On the Promises and Challenges of Multimodal Foundation Models for Geographical, Environmental, Agricultural, and Urban Planning Applications},
  author={Tan, Chenjiao and Cao, Qian and Li, Yiwei and Zhang, Jielu and Yang, Xiao and Zhao, Huaqin and Wu, Zihao and Liu, Zhengliang and Yang, Hao and Wu, Nemin and others},
  journal={arXiv preprint arXiv:2312.17016},
  year={2023}
}

@article{nelson1998time,
  title={Time series analysis using autoregressive integrated moving average (ARIMA) models},
  author={Nelson, Brian K},
  journal={Academic emergency medicine},
  volume={5},
  number={7},
  pages={739--744},
  year={1998},
  publisher={Wiley Online Library}
}

@article{lopez2023can,
  title={Can chatgpt forecast stock price movements? return predictability and large language models},
  author={Lopez-Lira, Alejandro and Tang, Yuehua},
  journal={arXiv preprint arXiv:2304.07619},
  year={2023}
}

@article{makridakis2023large,
  title={Large language models: Their success and impact},
  author={Makridakis, Spyros and Petropoulos, Fotios and Kang, Yanfei},
  journal={Forecasting},
  volume={5},
  number={3},
  pages={536--549},
  year={2023},
  publisher={MDPI}
}

@article{yang2021fact,
  title={Fact Check: Analyzing Financial Events from Multilingual News Sources},
  author={Yang, Linyi and Ng, Tin Lok James and Smyth, Barry and Dong, Ruihai},
  journal={arXiv preprint arXiv:2106.15221},
  year={2021}
}

@inproceedings{visintin2023leveraging,
  title={Leveraging Social Networks for Mergers and Acquisitions Forecasting},
  author={Visintin, Alessandro and Conti, Mauro},
  booktitle={International Conference on Web Information Systems Engineering},
  pages={144--159},
  year={2023},
  organization={Springer}
}

@article{chen2018models,
  title={Models for Predicting Business Bankruptcies and Their Application to Banking and Financial Regulation},
  author={Chen, James Ming},
  journal={Penn St. L. Rev.},
  volume={123},
  pages={735},
  year={2018},
  publisher={HeinOnline}
}

@inproceedings{ahmadi2018towards,
  title={Towards bankruptcy prediction: Deep sentiment mining to detect financial distress from business management reports},
  author={Ahmadi, Zahra and Martens, Peter and Koch, Christopher and Gottron, Thomas and Kramer, Stefan},
  booktitle={2018 IEEE 5th International Conference on Data Science and Advanced Analytics (DSAA)},
  pages={293--302},
  year={2018},
  organization={IEEE}
}

@inproceedings{kim2021corporate,
  title={Corporate bankruptcy prediction with domain-adapted BERT},
  author={Kim, Alex G and Yoon, Sangwon},
  booktitle={EMNLP 2021, 3rd Workshop on ECONLP},
  year={2021}
}

@inproceedings{lopez2016paysim,
  title={PaySim: A financial mobile money simulator for fraud detection},
  author={Lopez-Rojas, Edgar and Elmir, Ahmad and Axelsson, Stefan},
  booktitle={28th European Modeling and Simulation Symposium, EMSS, Larnaca},
  pages={249--255},
  year={2016},
  organization={Dime University of Genoa}
}

@inproceedings{roy2018deep,
  title={Deep learning detecting fraud in credit card transactions},
  author={Roy, Abhimanyu and Sun, Jingyi and Mahoney, Robert and Alonzi, Loreto and Adams, Stephen and Beling, Peter},
  booktitle={2018 systems and information engineering design symposium (SIEDS)},
  pages={129--134},
  year={2018},
  organization={IEEE}
}

@article{luo2023ai,
  title={AI-powered Fraud Detection in Decentralized Finance: A Project Life Cycle Perspective},
  author={Luo, Bingqiao and Zhang, Zhen and Wang, Qian and Ke, Anli and Lu, Shengliang and He, Bingsheng},
  journal={arXiv preprint arXiv:2308.15992},
  year={2023}
}

@article{feng2023empowering,
  title={Empowering many, biasing a few: Generalist credit scoring through large language models},
  author={Feng, Duanyu and Dai, Yongfu and Huang, Jimin and Zhang, Yifang and Xie, Qianqian and Han, Weiguang and Lopez-Lira, Alejandro and Wang, Hao},
  journal={arXiv preprint arXiv:2310.00566},
  year={2023}
}

@book{brigo2006interest,
  title={Interest rate models-theory and practice: with smile, inflation and credit},
  author={Brigo, Damiano and Mercurio, Fabio},
  volume={2},
  year={2006},
  publisher={Springer}
}

@book{lee2019financial,
  title={Financial econometrics, mathematics and statistics},
  author={Lee, Cheng-Few and Chen, Hong-Yi and Lee, John},
  year={2019},
  publisher={Springer}
}

@article{buehler2019deep,
  title={Deep hedging},
  author={Buehler, Hans and Gonon, Lukas and Teichmann, Josef and Wood, Ben},
  journal={Quantitative Finance},
  volume={19},
  number={8},
  pages={1271--1291},
  year={2019},
  publisher={Taylor \& Francis}
}

@article{cont2001empirical,
  title={Empirical properties of asset returns: stylized facts and statistical issues},
  author={Cont, Rama},
  journal={Quantitative finance},
  volume={1},
  number={2},
  pages={223},
  year={2001},
  publisher={IOP Publishing}
}

@article{araci2019finbert,
  title={Finbert: Financial sentiment analysis with pre-trained language models},
  author={Araci, Dogu},
  journal={arXiv preprint arXiv:1908.10063},
  year={2019}
}

@article{shah2022flue,
  title={When flue meets flang: Benchmarks and large pre-trained language model for financial domain},
  author={Shah, Raj Sanjay and Chawla, Kunal and Eidnani, Dheeraj and Shah, Agam and Du, Wendi and Chava, Sudheer and Raman, Natraj and Smiley, Charese and Chen, Jiaao and Yang, Diyi},
  journal={arXiv preprint arXiv:2211.00083},
  year={2022}
}

@inproceedings{deng2023llms,
  title={What do llms know about financial markets? a case study on reddit market sentiment analysis},
  author={Deng, Xiang and Bashlovkina, Vasilisa and Han, Feng and Baumgartner, Simon and Bendersky, Michael},
  booktitle={Companion Proceedings of the ACM Web Conference 2023},
  pages={107--110},
  year={2023}
}

@article{gupta2023gpt,
  title={GPT-InvestAR: Enhancing Stock Investment Strategies through Annual Report Analysis with Large Language Models},
  author={Gupta, Udit},
  journal={arXiv preprint arXiv:2309.03079},
  year={2023}
}

@article{wu2023brief,
  title={A brief overview of ChatGPT: The history, status quo and potential future development},
  author={Wu, Tianyu and He, Shizhu and Liu, Jingping and Sun, Siqi and Liu, Kang and Han, Qing-Long and Tang, Yang},
  journal={IEEE/CAA Journal of Automatica Sinica},
  volume={10},
  number={5},
  pages={1122--1136},
  year={2023},
  publisher={IEEE}
}

@article{malo2014good,
  title={Good debt or bad debt: Detecting semantic orientations in economic texts},
  author={Malo, Pekka and Sinha, Ankur and Korhonen, Pekka and Wallenius, Jyrki and Takala, Pyry},
  journal={Journal of the Association for Information Science and Technology},
  volume={65},
  number={4},
  pages={782--796},
  year={2014},
  publisher={Wiley Online Library}
}

@inproceedings{maia201818,
  title={Www'18 open challenge: financial opinion mining and question answering},
  author={Maia, Macedo and Handschuh, Siegfried and Freitas, Andr{\'e} and Davis, Brian and McDermott, Ross and Zarrouk, Manel and Balahur, Alexandra},
  booktitle={Companion proceedings of the the web conference 2018},
  pages={1941--1942},
  year={2018}
}

@inproceedings{alvarado2015domain,
  title={Domain adaption of named entity recognition to support credit risk assessment},
  author={Alvarado, Julio Cesar Salinas and Verspoor, Karin and Baldwin, Timothy},
  booktitle={Proceedings of the Australasian Language Technology Association Workshop 2015},
  pages={84--90},
  year={2015}
}

@article{chen2021finqa,
  title={Finqa: A dataset of numerical reasoning over financial data},
  author={Chen, Zhiyu and Chen, Wenhu and Smiley, Charese and Shah, Sameena and Borova, Iana and Langdon, Dylan and Moussa, Reema and Beane, Matt and Huang, Ting-Hao and Routledge, Bryan and others},
  journal={arXiv preprint arXiv:2109.00122},
  year={2021}
}

@article{chen2022convfinqa,
  title={Convfinqa: Exploring the chain of numerical reasoning in conversational finance question answering},
  author={Chen, Zhiyu and Li, Shiyang and Smiley, Charese and Ma, Zhiqiang and Shah, Sameena and Wang, William Yang},
  journal={arXiv preprint arXiv:2210.03849},
  year={2022}
}

@article{gage1994new,
  title={A new algorithm for data compression},
  author={Gage, Philip},
  journal={C Users Journal},
  volume={12},
  number={2},
  pages={23--38},
  year={1994},
  publisher={McPherson, KS: R \& D Publications, c1987-1994.}
}

@article{liu2024understanding,
  title={Understanding LLMs: A Comprehensive Overview from Training to Inference},
  author={Liu, Yiheng and He, Hao and Han, Tianle and Zhang, Xu and Liu, Mengyuan and Tian, Jiaming and Zhang, Yutong and Wang, Jiaqi and Gao, Xiaohui and Zhong, Tianyang and others},
  journal={arXiv preprint arXiv:2401.02038},
  year={2024}
}

@inproceedings{zhang2023enhancing,
  title={Enhancing financial sentiment analysis via retrieval augmented large language models},
  author={Zhang, Boyu and Yang, Hongyang and Zhou, Tianyu and Ali Babar, Muhammad and Liu, Xiao-Yang},
  booktitle={Proceedings of the Fourth ACM International Conference on AI in Finance},
  pages={349--356},
  year={2023}
}

@article{radford2019language,
  title={Language models are unsupervised multitask learners},
  author={Radford, Alec and Wu, Jeffrey and Child, Rewon and Luan, David and Amodei, Dario and Sutskever, Ilya and others},
  journal={OpenAI blog},
  volume={1},
  number={8},
  pages={9},
  year={2019}
}

@article{sun2024trustllm,
  title={TrustLLM: Trustworthiness in Large Language Models},
  author={Sun, Lichao and Huang, Yue and Wang, Haoran and Wu, Siyuan and Zhang, Qihui and Gao, Chujie and Huang, Yixin and Lyu, Wenhan and Zhang, Yixuan and Li, Xiner and others},
  journal={arXiv preprint arXiv:2401.05561},
  year={2024}
}

@article{wang2024large,
  title={Large Language Models for Robotics: Opportunities, Challenges, and Perspectives},
  author={Wang, Jiaqi and Wu, Zihao and Li, Yiwei and Jiang, Hanqi and Shu, Peng and Shi, Enze and Hu, Huawen and Ma, Chong and Liu, Yiheng and Wang, Xuhui and others},
  journal={arXiv preprint arXiv:2401.04334},
  year={2024}
}

@article{liu2023radiology,
  title={Radiology-GPT: A Large Language Model for Radiology},
  author={Liu, Zhengliang and Zhong, Aoxiao and Li, Yiwei and Yang, Longtao and Ju, Chao and Wu, Zihao and Ma, Chong and Shu, Peng and Chen, Cheng and Kim, Sekeun and others},
  journal={arXiv preprint arXiv:2306.08666},
  year={2023}
}

@article{liu2023context,
  title={Context matters: A strategy to pre-train language model for science education},
  author={Liu, Zhengliang and He, Xinyu and Liu, Lei and Liu, Tianming and Zhai, Xiaoming},
  journal={arXiv preprint arXiv:2301.12031},
  year={2023}
}

\end{document}